\documentclass[10pt,twocolumn,letterpaper]{article}
\usepackage[accsupp]{axessibility}
\usepackage[pagenumbers]{cvpr}

\usepackage[final]{microtype}

\newcommand{\ours}{LightMover\xspace}

\usepackage[dvipsnames]{xcolor}
\usepackage{colortbl}
\usepackage{multirow}
\usepackage{pifont}
\newcommand{\cmark}{\ding{51}}
\newcommand{\xmark}{\ding{55}}
\usepackage{tcolorbox}

\definecolor{cvprblue}{rgb}{0.21,0.49,0.74}
\usepackage[pagebackref,breaklinks,colorlinks,allcolors=cvprblue]{hyperref}
\newcommand\blfootnote[1]{
    \begingroup
    \renewcommand\thefootnote{}\footnote{#1}
    \addtocounter{footnote}{-1}
    \endgroup
}

\title{\ours: Generative Light Movement with Color and Intensity Controls}

\author{%
  Gengze Zhou$^{1,*}$\quad
  Tianyu Wang$^{2,\dagger}$\quad
  Soo Ye Kim$^2$\quad
  Zhixin Shu$^2$\quad
  Xin Yu$^3$\quad \\
  Yannick Hold-Geoffroy$^2$\quad
  Sumit Chaturvedi$^4$\quad
  Qi Wu$^1$\quad
  Zhe Lin$^2$\quad
  Scott Cohen$^2$ \\
  $^1$AIML, Adelaide University\quad
  $^2$Adobe Research \quad
  $^3$University of Hong Kong \quad
  $^4$Yale University\\
  \tt\small {gengze.zhou@adelaide.edu.au, stevewong.cv@gmail.com}\\
  {\small \url{https://gengzezhou.github.io/LightMover/}}
}

\begin{document}
\twocolumn[{%
\renewcommand\twocolumn[1][]{#1}%
\maketitle

\begin{center}
    \centering
    \captionsetup{type=figure}
    \includegraphics[width=0.91\textwidth]{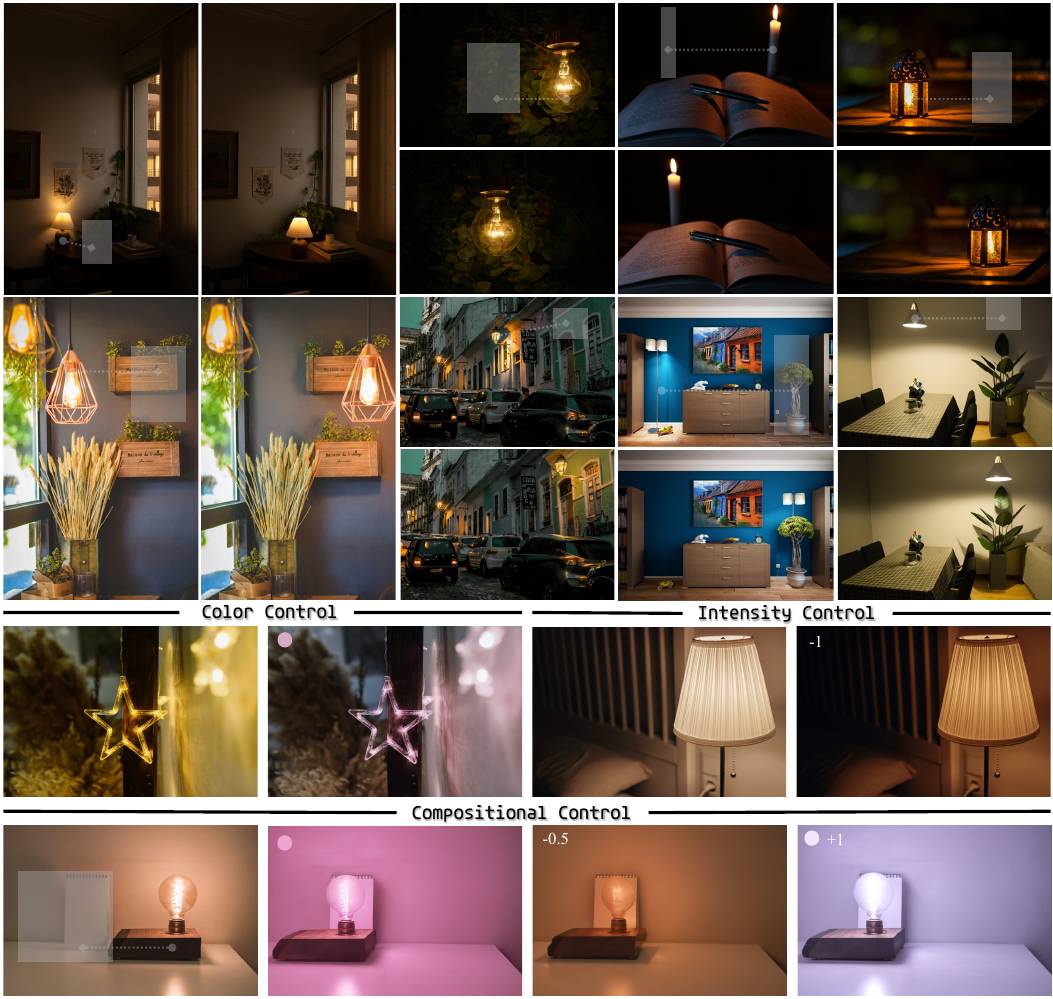}
    \vspace{-0.5em}
    \captionof{figure}{\textbf{Results of light source movement.} \ours demonstrates robust control over light movement, color, and intensity in diverse illumination scenarios. It maintains global radiometric balance, reconstructs occluded illumination, and infers plausible light–object interactions under varying geometry and materials. Beyond single-attribute control, \ours supports \textbf{compositional manipulation} that simultaneously adjusts light position, color, and intensity. These results highlight the potential of our 2.5D learning paradigm to achieve physically coherent lighting reasoning within a purely 2D generative framework.}

    \label{fig:teaser}
\end{center}%
}]

\blfootnote{* Work done during an internship at Adobe.}
\blfootnote{$\dagger$ Corresponding author.}

\begin{abstract}
We present \ours, a framework for controllable light manipulation in single images that leverages video diffusion priors to produce physically plausible illumination changes without re-rendering the scene. We formulate light editing as a sequence-to-sequence prediction problem in visual token space: given an image and light-control tokens, the model adjusts light position, color, and intensity together with resulting reflections, shadows, and falloff from a single view. This unified treatment of spatial (movement) and appearance (color, intensity) controls improves both manipulation and illumination understanding.
We further introduce an adaptive token-pruning mechanism that preserves spatially informative tokens while compactly encoding non-spatial attributes, reducing control sequence length by $41\%$ while maintaining editing fidelity. To train our framework, we construct a scalable rendering pipeline that generates large numbers of image pairs across varied light positions, colors, and intensities while keeping the scene content consistent with the original image. \ours enables precise, independent control over light position, color, and intensity, and achieves high PSNR and strong semantic consistency (DINO, CLIP) across different tasks.
\end{abstract}    
\section{Introduction}

Imagine shopping online for a new floor lamp. Product photos reveal its appearance, but not what truly matters: how the lamp would change the mood and illumination of your own living room. With only a single image of a living room, a user naturally wants to insert a lamp, drag it across the room, warm its color, dim its brightness, and observe how these adjustments reshape shadows, reflections, and ambience. Achieving such light-centric editing in a single real image is extremely challenging as illumination interacts globally with geometry, materials, and occlusion, while the user expects precise, causal control.

Existing methods fall into two categories. Inverse-rendering approaches~\cite{li20_inverse_indoor,sengupta19_nir,li22_indoor_light,wang25_materialist,wu25_pbrnerf} reconstruct dense geometry, materials, and lighting before re-rendering the scene. While physically grounded, these pipelines are highly ill-posed from a single image, computationally expensive, and do not offer intuitive controls, such as moving or recoloring individual lights interactively. Diffusion-based editors (e.g., LightLab~\cite{magar2025lightlab}) can manipulate tone and ambient illumination but do not model spatial light movement, while object-centric diffusion models (e.g., ObjectMover~\cite{yu25_objectmover}) rely on video priors to move objects but treat illumination only implicitly, and thus fail to propagate the correct shadows, colors, and falloff when moving a light source.
More recently, general generative image editing models (SDEdit \cite{meng2021sdedit}, DiffEdit \cite{couairon23_diffedit}, InstructPix2Pix \cite{brooks23_instructpix2pix}, Nano-Banana \cite{google25_gemini_flash_image}, Qwen-Image(-Edit) \cite{wu25_qwenimage, qwen25_image_edit}) support powerful text- or mask-driven semantic edits; however, they lack an explicit light-centric parameterization and thus struggle to provide precise, physically plausible control over individual lights.

In this paper, we introduce LightMover, a video-diffusion framework for explicit, parametric illumination control from a single image. Our key idea is to extend the ObjectMover \cite{yu25_objectmover} token formulation with illumination-specific control tokens for color and intensity, embedded using a multi-signal positional encoding (spatial, temporal, condition-type, and role). This design yields a unified model that supports both object edits (movement, insertion, removal) and illumination edits within the same backbone, avoiding task-specific adapters. A naive extension of this formulation is to encode color and intensity as extra “frames”, but this dramatically increases the token count and thereby limits the model’s efficiency and resolution. To avoid control tokens inflating the sequence length, we introduce an adaptive token-pruning strategy: spatial controls (e.g., movement maps) retain fine-grained tokens in small, localized regions and are downsampled elsewhere, while non-spatial controls (color, intensity) are represented with learnable compression ratios. This reduces the length of the control token by $41\%$ while maintaining accurate spatial and illumination control.

To support training, we build a scalable physically-based rendering pipeline that generates matched image pairs varying in light location, spectrum, and intensity. These pairs enable the model to learn the causal effects of illumination: how shadows shift, how reflections brighten, and how indirect lighting propagates, to unlock capabilities that video priors alone cannot provide.
Moreover, we find that training these illumination controls jointly in a multi-task manner produces strong mutual reinforcement: positional, color, and intensity signals regularize one another, improving accuracy and stability across all light-related tasks.

In summary, our contributions are threefold:
\begin{itemize}
    \item We introduce a unified diffusion-based framework that formulates light control as a sequence-to-sequence task, enabling fine-grained manipulation of illumination.
    \item We propose an adaptive token-pruning mechanism that improves efficiency by compactly encoding non-spatial attributes while preserving spatial precision.
    \item We develop a scalable data generation pipeline with physically accurate rendering and a multi-task training strategy for light insertion, removal, and movement, resulting in state-of-the-art photorealistic and controllable lighting manipulation.
\end{itemize}
\section{Related Works}

\noindent\textbf{Object Movement.}
Diffusion-based inpainting methods~\citep{avrahami2022blended, suvorov2021resolution, lugmayr2022repaint, meng2021sdedit, saharia2022palette} represent the mainstream approach for generative removal, while insertion frameworks~\citep{lu2023tficon, zhang2023controlcom, anydoor, paint_by_example, objectstitch, imprint} extend them to conditionally synthesize new objects under identity constraints. However, inpainting inherently limits the handling of global illumination cues, as expanding masks often destroy scene consistency and fail to reconstruct shadows or reflections. Shadow-specific methods~\citep{guo2023shadowdiffusion, wang2020instance} partially alleviate this but rely on precise masks and cannot generalize to broader lighting effects. Mask-free approaches such as ObjectDrop~\citep{objectdrop} decouple removal and insertion into two steps, often leading to misaligned perspective or inconsistent lighting, while scene-level insertion~\citep{think_outside} allows broader context editing but lacks true movement capability.
Recent progress in generative movement seeks to reposition objects directly through single-pass diffusion. Control paradigms include drag-based manipulation~\citep{mou2023dragondiffusion, shi2024dragdiffusion, avrahami2024diffuhaul, mou2024diffeditor}, motion-field guidance~\citep{geng2024motion}, 3D-aware conditioning~\citep{3dit, diff_handle, yenphraphai2024image, bhat2023loosecontrol}, and video diffusion-based sequence modeling~\citep{yu25_objectmover}.

\vspace{0.05in}\noindent\textbf{Relighting and Light Control.}
Image-based light editing has been widely explored in flash photography, addressing white balancing~\citep{hui2016whitebalance}, bounce-flash capture~\citep{murmann2016bounceflash}, and post-capture flash control~\citep{maralan2023flashintrinsics}. Murmann et al.~\citep{murmann2019multiillumination} collected a multi-illumination dataset using a motorized flash rig and trained a U-Net~\citep{ronneberger2015unet} to predict directional lighting. Hui et al.~\citep{hui2017spectra} used flash/no-flash pairs to disentangle illuminants based on spectral cues, enabling control over individual light sources. Aksoy et al.~\citep{aksoy2018flashambient} curated a large flash/no-flash dataset and trained a Pix2Pix model~\citep{isola2017pix2pix} to separate ambient and flash components.
Inverse-rendering pipelines~\citep{debevec2000acquiring,lensch2003image} reconstruct geometry and materials for explicit lighting control, but are computationally expensive and ill-posed from a single image; recent work~\citep{careaga2025physically,zeng2024rgb,liang2025diffusion,xing2025luminet,zhang2025scaling} incorporates learned priors to improve scalability while still relying on intrinsic decomposition or dense video supervision.
We follow a similar data-driven paradigm but in a more general setting where only a single RGB image is available. Unlike prior flash-specific approaches, we draw inspiration from LightLab~\citep{magar2025lightlab}, which leverages diffusion models for explicit light-source control and turn-on/off illumination synthesis. Note that LightLab does not support light source movement.

\vspace{0.05in}\noindent\textbf{General Image Editing.}
Beyond task-specific pipelines, diffusion-based editors provide broad semantic and structural manipulation capabilities. Classical methods such as SDEdit~\citep{meng2021sdedit} and DiffEdit~\citep{couairon23_diffedit} enable mask-driven or noise-based refinements, while instruction-tuned models like InstructPix2Pix~\citep{brooks23_instructpix2pix} support natural-language edits. Recent LLM-powered systems, including Nano-Banana~\citep{Gemini2.5Flash2025} and Qwen-Image(-Edit)~\citep{wu25_qwenimage, qwen25_image_edit}, further enhance multimodal reasoning and editing robustness. However, these general-purpose editors lack an explicit light-centric formulation and therefore struggle to enforce physically plausible illumination, realistic shadow propagation, and precise spatial control needed for light movement and fine-grained relighting.

\section{Method}

\subsection{LightMover}

\begin{figure*}
    \centering
    \includegraphics[width=0.98\linewidth]{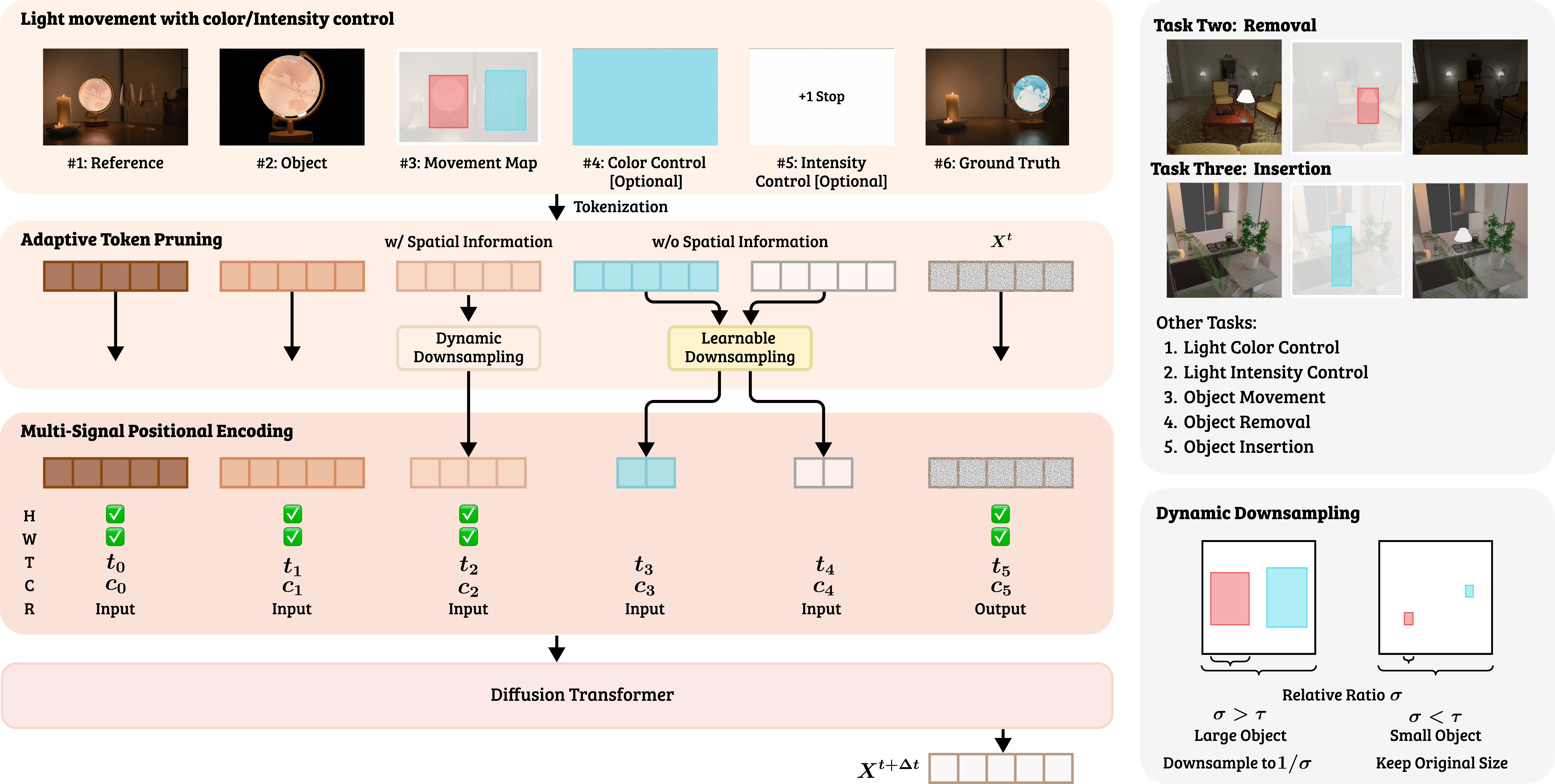}
    \caption{\textbf{Overview of \ours.} Left: our sequence-to-sequence formulation encodes a reference image, object crop, movement map, and optional color/intensity controls as input frames. Right: additional supported tasks, including light removal and insertion.}
    \label{figs:model}
\end{figure*}

\subsubsection{Repurposing Video Diffusion Models}

We follow ObjectMover~\cite{yu25_objectmover} to repurpose a pre-trained image-to-video diffusion transformer model~\cite{transformer, dit}, similar in design to Sora~\cite{sora}, within a \textbf{sequence-to-sequence generation framework} that maintains compatibility with the original video input interface while targeting single-frame image synthesis. As illustrated in \cref{figs:model}, we formulate all input conditions as pseudo video frames arranged sequentially. Each frame is encoded by a VAE into latent tokens and jointly processed by the diffusion transformer.

Specifically, the sequence elements are defined as follows:  
(1) \textbf{Reference Image} $I_\text{ref}$: the original input image for relighting or motion editing.  
(2) \textbf{Object Frame} $I_\text{obj}$: the cropped and resized target object to be manipulated.  
(3) \textbf{Movement Map} $I_\text{move}$: the spatial control signal indicating object displacement. It encodes two bounding boxes: source ($M_\text{src}$) and target ($M_\text{tgt}$) across separate RGB channels, where the \textit{R} channel denotes the region to be removed (source), and the \textit{GB} channels indicate the target region to which the object will move.  
(4) \textbf{Color Control} $I_\text{color}$: a global conditioning frame specifying the desired lighting hue or color temperature.  
(5) \textbf{Intensity Control} $I_\text{intensity}$: a frame that encodes the variation in light exposure, expressed in photographic \textit{stops}. A single \textit{stop} corresponds to a twofold increase or decrease in the amount of light reaching the sensor and serves as a standard unit for quantifying exposure changes in imaging research~\cite{dolhasz2020learning}. The corresponding illumination gain, representing the multiplicative change in radiometric intensity induced by an exposure adjustment of $S_\mathrm{EV}$ \textit{stops}, is defined as:
\begin{equation}
    G_\mathrm{illum} = 2^{\,S_\mathrm{EV}},
    \label{eq:illum_gain}
\end{equation}
where $G_\mathrm{illum}$ is the illumination gain (a dimensionless ratio of radiometric intensity), and $S_\mathrm{EV}$ denotes the exposure adjustment in \textit{stops} (Exposure Value, EV). An increment of $+1$\,EV doubles the illumination, whereas a decrement of $-1$\,EV halves it.
(6) \textbf{Output Frame} $X^t$: the noisy latent frame to be denoised at timestep $t$.
All frames are encoded by the video VAE into latent tokens before being fed to the diffusion model.

During inference, generation begins from Gaussian noise $X^0 \!\sim\! \mathcal{N}(0,1)$.  
At each denoising step $t$, the conditional sequence $S^t$ is formed as:
\begin{equation}
S^t = [I_\text{ref}, I_\text{obj}, I_\text{move}, I_\text{color}, I_\text{intensity}, X^t],
\end{equation}
and iteratively refined by the model $v_\theta$ to predict the clean target image $X^1$.  
The model outputs an updated sequence $S^{t+\Delta t}$, and prior to the next step, all condition frames are reset to their original latent states for stability.

During training, the control signals can be flexibly combined to synthesize compositional editing tasks, or selectively dropped to enable single-attribute editing. Training follows a \textbf{flow-matching objective}~\cite{flow_matching,instaflow}.  
Noisy inputs are generated by linear interpolation:
\begin{equation}
X^t = tX^1 + (1 - t)X^0, \quad X^0 \!\sim\! \mathcal{N}(0,1),
\end{equation}
where the model predicts the instantaneous velocity:
\begin{equation}
V^t = \frac{dX^t}{dt} = X^1 - X^0.
\end{equation}
The loss is expressed as:
\begin{equation}
\mathcal{L} = \mathbb{E}_{t,X^0,X^1}\!\left[\left\|v(S^t, t; \theta)_{[6]} - V^t\right\|^2\right],
\end{equation}
where $v_{[6]}$ denotes the output token of the target frame.

\subsubsection{Multi-Signal Positional Encoding}
To ensure the diffusion transformer interprets each input frame according to its spatial meaning and control semantics, we augment the model with a specialized positional encoding method.
To jointly encode spatial, temporal, and conditional context within the diffusion transformer, we introduce a \textbf{Multi-Signal Positional Encoding (MSPE)} mechanism that extends beyond standard rotary positional embeddings (RoPE)~\cite{su2024roformer}.  
MSPE integrates four orthogonal positional subspaces: 
(1) \textbf{Spatial Encoding ($W,H$):} Each latent patch token is assigned a 2D sine-cosine positional encoding according to its horizontal ($W$) and vertical ($H$) coordinates, preserving spatial structure within each frame.
(2) \textbf{Temporal Encoding ($T$):} Tokens receive temporal indices reflecting the sequence order, enabling temporal consistency across diffusion steps.
(3) \textbf{Condition-Type Encoding ($C$):} Tokens are annotated with discrete condition identifiers (\textit{reference, object, movement, color, intensity, output}), to distinguish different conditioning modalities.
(4) \textbf{Frame-Role Encoding ($R$):} To differentiate \textit{input} conditions from the \textit{output} frame, a binary encoding is applied, marking whether a token contributes to conditioning or prediction.

These four components are projected into the transformer embedding space and additively combined, followed by a rotation-aware modulation analogous to RoPE to preserve relative phase relationships.  
For variable-length sequences and spatial resolutions, we employ \textbf{NTK-aware interpolation}~\cite{peng2023yarn} to dynamically adapt positional frequencies dynamically.  
This unified encoding enables the model to reason jointly over spatial alignment and condition interdependence, supporting precise and disentangled control over object motion, color, and light intensity.

\subsubsection{Adaptive Token Pruning}

In our formulation, adding additional control signals inevitably increases the sequence length. For instance, a $512{\times}512$ image encoded by a video VAE with a downsampling factor of $32$ yields $256$ latent tokens per frame. However, certain attribute frames, such as those for color or intensity lack explicit spatial structure, making it unnecessary to preserve their full two-dimensional token layout.  

To improve computational efficiency while maintaining generation fidelity, we introduce an \textbf{adaptive token pruning} mechanism that dynamically adjusts the number of latent tokens for each condition. Two complementary strategies are employed:
(1) \textbf{Spatially-Aware Pruning:} For control signals with explicit spatial information, such as the movement map $I_\text{move}$, we compute the area ratio of the bounding box relative to the full frame. If the ratio is below a threshold $\tau = 0.2$ (i.e., corresponding to a small object), we retain the full-resolution latent map; otherwise, we downsample the latent tokens proportionally to the area ratio, effectively reducing redundant background information.
(2) \textbf{Learnable Downsampling for Non-Spatial Signals:} For non-spatial control frames such as color and intensity, we introduce learnable downsampling ratios. The number of retained tokens for the color frame ($I_\text{color}$) and intensity frame ($I_\text{intensity}$) are both optimized jointly with the diffusion model, balancing efficiency and representational fidelity.

This adaptive pruning mechanism minimizes redundant computation across spatially sparse or low-dimensional conditions while preserving fine-grained control for localized edits such as object movement or illumination adjustment. In practice, it reduces the average control sequence length by \textbf{41\%} in \ours, significantly improving efficiency without compromising generation quality.

\subsection{Data Generation Pipeline}
\label{sec:datagen}

\begin{figure}
    \centering
    \includegraphics[width=1\linewidth]{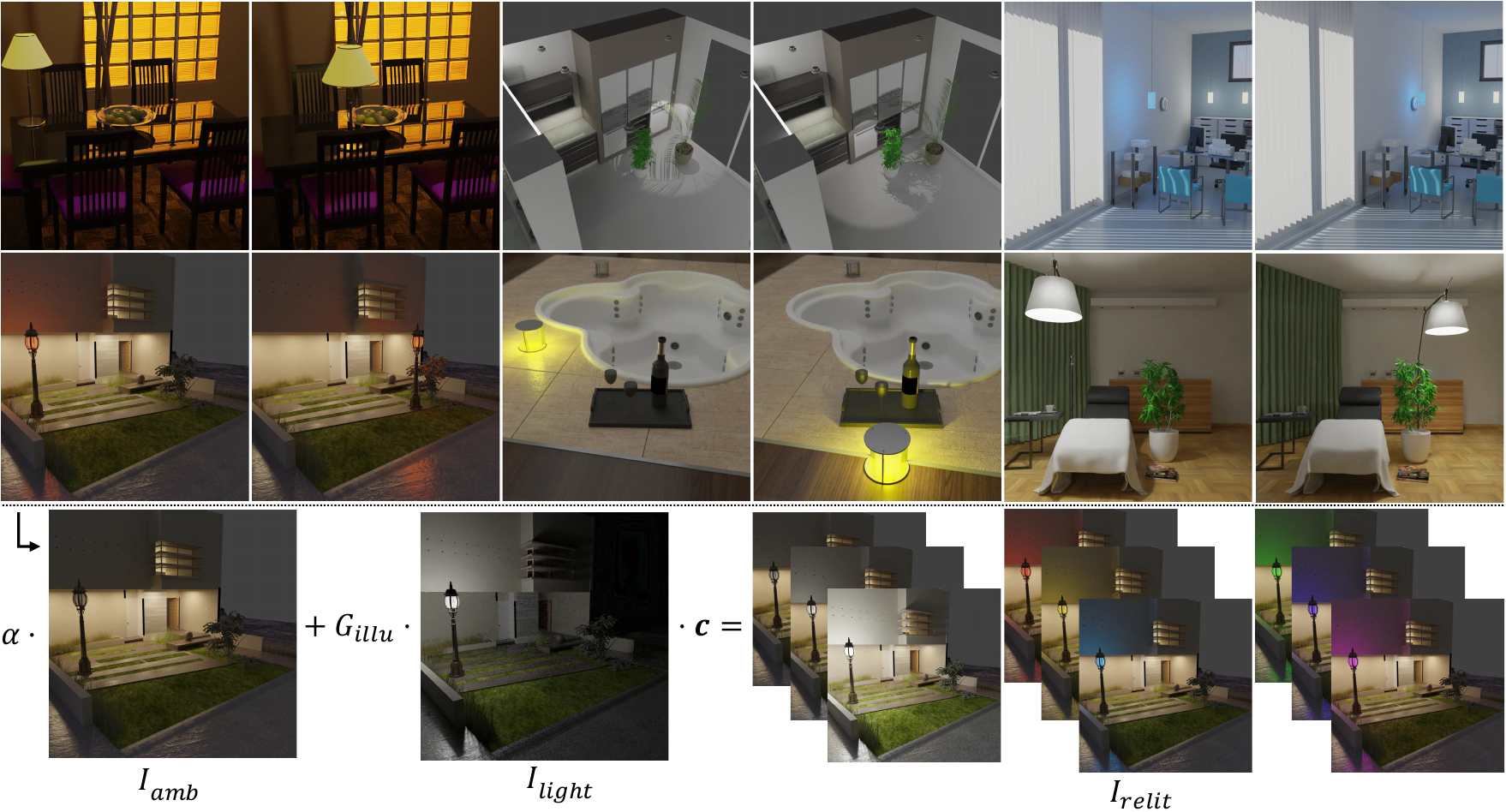}
    \caption{\textbf{Illustration of training samples.} Top: our synthetic data varies from scenes, objects, and lighting conditions; Bottom: the image relighting post-processing during training.}
    \label{fig:data}
\end{figure}

To train a model capable of accurate and physically consistent light manipulation, we combine \textbf{real captured data} with a \textbf{large-scale synthetic corpus}. Real-world photographs provide natural appearance and material diversity, while synthetic data enable systematic variation of lighting parameters and scalable supervision across light movement, color, and intensity. Together, these two sources form a comprehensive dataset that supports learning both visual realism and physical consistency.

\subsubsection{Synthetic Rendering Data Pipeline}
\noindent\textbf{Scene and Lighting Setup.}
We use 25 artist-designed indoor environments created in \textit{Blender}~\cite{Blender2018}, from which we systematically render scenes under varying lighting and object configurations. To diversify scenes, we randomly sample plausible \textit{fixture placements} (e.g., ceiling lamps, wall sconces, desk lights), vary HDRI environment maps, and adjust ambient-to-direct light ratios.
For object diversity, we retrieve 100 light source assets from \textit{Objectverse-XL}~\cite{objaverseXL}, selected by CLIP similarity to the query ``light,'' and preprocess them in Blender to normalize scale, emissive materials, and mounting anchors. In each scene, a selected light is animated along a smooth trajectory, with ten virtual cameras capturing multi-view motion. Consecutive frames form light movement pairs.
This pipeline generates about 32{,}000 combinations of data pairs with various light sources, camera views, and illumination conditions, as illustrated in Figure~\ref{fig:data}.

\noindent\textbf{Physically Disentangled Rendering.}
During training, additional perturbations of light intensity, hue, and ambient tone are applied in post-processing, creating effectively unbounded variations.
Each rendered frame is decomposed into two physically disentangled components: ambient base image $I_{\text{amb}}$ and direct light contribution $I_{\text{light}}$.
These are rendered independently under a Monte Carlo path-tracing setup and later composited in linear RGB space as:
\begin{equation}
    I_{\text{lin}} = I_{\text{amb}} + I_{\text{light}} .
\end{equation}

\noindent\textbf{Parametric Light Control.}
To synthesize illumination variations, we adjust the relative strength and color of controllable light sources. Let
$\alpha \in [0,1]$ denote the ambient scaling factor,
$G_\text{illum} \in [0,1]$ the \textit{target-light intensity gain}, and
$\mathbf{c}_t \in \mathbb{R}^3$ the \textit{desired RGB tint} in linear color space.
The relit image is computed as:
\begin{equation}
    I_{\text{relit}}(\alpha, G_\text{illum}, \mathbf{c}_t)
    = \alpha\, I_{\text{amb}}
    + G_\text{illum}\, I_{\text{light}} \odot \mathbf{c}_t ,
    \label{eq:relighting}
\end{equation}
where $\odot$ denotes element-wise multiplication across RGB channels.
During rendering, each light’s emission color is set to pure white and unit intensity, so the target tint color could be directly applied in post-processing.

\noindent\textbf{Tone Mapping.}
Since physically based rendering operates in linear RGB, pixel intensities may exceed displayable ranges due to rare high-energy samples.
We stabilize the dynamic range via percentile-based normalization followed by sRGB tone mapping.
Let $E_{\max}$ be the 99.95th percentile of pixel luminance over 1,024 random samples.
The tone-mapped image is computed as:
\begin{equation}
    I_{\text{srgb}} = 
    \mathrm{clip}\!\left(
    \frac{I_{\text{lin}}}{E_{\max}}
    \right)^{1/2.2},
    \label{eq:tonemapping}
\end{equation}
where the exponent $1/2.2$ approximates the standard sRGB gamma curve and $\mathrm{clip}(\cdot)$ bounds values to $[0,1]$.

\begin{figure*}[h]
    \centering
    \includegraphics[width=1\linewidth]{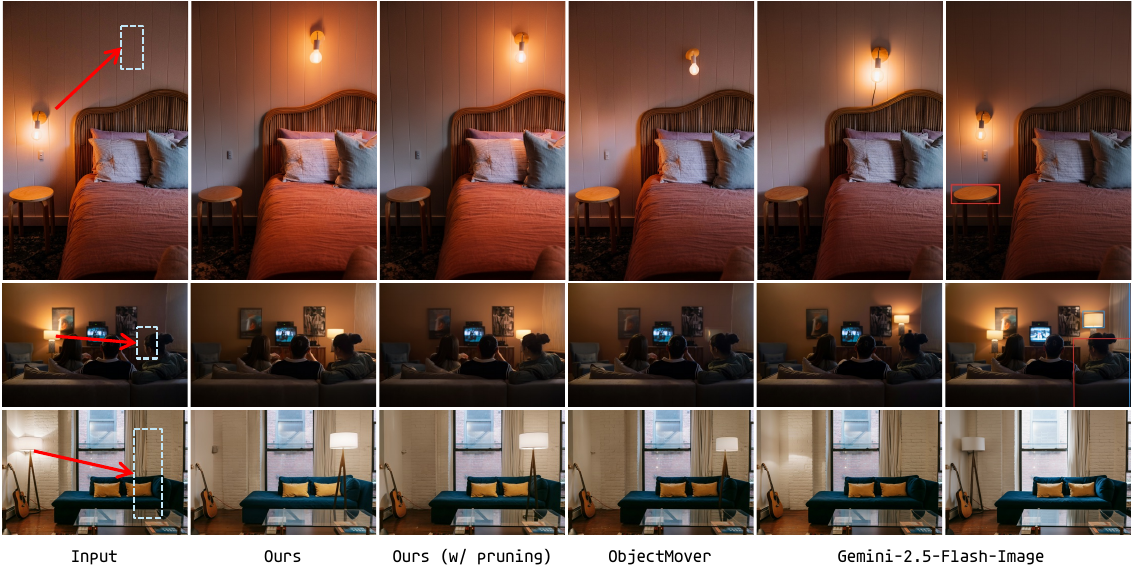}
    \caption{\textbf{Comparison of light movement results.} The blue bounding box in the input indicates the target light location. For \textit{Gemini-2.5-Flash-Image}, the first column shows the two-step modular editing setup, and the second column shows the one-step editing result. Compared with \textit{ObjectMover}, \ours accurately handles light propagation and its interaction with the environment. Compared with \textit{Gemini-2.5-Flash-Image}, \ours provides more precise control and produces more coherent and photometrically consistent relighting.}
    \label{fig:comp}
\end{figure*}
\subsubsection{Real Data Capture}

We capture a set of real-world image pairs using off-the-shelf mobile devices, a tripod, and synchronized triggering equipment. Each pair depicts the same scene in two conditions, where the only physical change is the location of a visible light source. The dataset covers 106 indoor scenes, each with 3-4 lighting variations, resulting in 360 high-resolution photographs. Beyond the pairs used for the light movement task, we also capture a background reference image for each scene, in which the light source is physically removed. These background images enable additional training objectives for light insertion and removal.

\section{Experiments}

We train \ours on a 5B-parameter video diffusion transformer. Training samples are drawn at mixed resolutions of 512×512 and 1024×1024 with a 1:1 ratio. The dataset combines synthetic and real data at a 10:1 ratio. Synthetic tasks are distributed across seven categories: (1) light movement, (2) object movement, (3) light color change, (4) light intensity change, (5) joint movement and color/intensity change, (6) light removal, and (7) light insertion, following a ratio of 6:3:3:3:1:1, respectively. We additionally apply light-illumination augmentation using the Physically Disentangled Rendering procedure described in Section~\ref{sec:datagen}, which dynamically varies ambient and direct light components during training. More implementation details and training hyperparameters will be provided in the supplementary file.

\noindent\textbf{Evaluation.}
We evaluate \ours quantitatively and qualitatively across two complementary benchmarks designed to assess controllable illumination editing: \textit{LightMove-A} for real captured data and \textit{LightMove-B} for synthetic scenes with ground-truth illumination variations.  

\noindent\textbf{LightMove-A (Real).}
\textit{LightMove-A} comprises 200 real-world image triplets captured by experienced photographers. Each set includes (1) a scene with the light source, (2) the same scene with the light moved. The dataset enables the evaluation of realistic light movement.

\noindent\textbf{LightMove-B (Synthetic).}
\textit{LightMove-B} is a synthetic benchmark to evaluate illumination control with known ground-truth light variations. 
These pairs are curated as described in Section~\ref{sec:datagen} and contain 2 held-out scenes with unseen light sources, objects, and materials.

\noindent\textbf{Metrics.}
For quantitative analysis, we use perceptual similarity metrics including DINO-Score~\cite{dino} and CLIP-Score~\cite{clip}, computed over the local regions affected by illumination changes. We additionally report PSNR on the full image to assess overall photometric consistency.

\noindent\textbf{Additional Evaluations.}
For qualitative evaluation, where explicit ground truth is not required, we use a collection of 50 diverse real images from Pexels~\cite{Pexels}. We manually annotate light-source regions and corresponding masks, and use these as references for visual comparison.

\subsection{Comparison to SoTA Models}
In this section, we extensively evaluate the light movement performance of \ours on various settings, as well as its general object movement ability.

\noindent\textbf{Light Movement.}
We evaluate the light movement performance of \ours on the \textit{LightMove-A} dataset for quantitative analysis and the Pexels dataset for qualitative comparison. Quantitative results are reported in Table~\ref{tab:light_movement}, and visual examples are shown in Figure~\ref{fig:comp}. We compare against two representative baseline streams:  
(1) \textit{LLM-powered text-to-image (T2I) models}, including Gemini-2.5-Flash-Image \cite{google25_gemini_flash_image} and Qwen-Image \cite{wu25_qwenimage}, which interpret text instructions through large language model integration. To ensure fair comparison, we evaluate these models under two settings:  
\textit{(a)} a one-step editing setup, where both the source object and target location are marked with bounding boxes and specified in the text prompt, and  
\textit{(b)} a two-step pipeline, where the model first removes the light source and then inserts it at the target position indicated by the bounding box. Details including the prompt templates are provided in the Supplementary Material.
(2) \textit{ObjectMover} \cite{yu25_objectmover}, 
an object movement model,
serving as a structural baseline.

\begin{table}[t]
\centering
\caption{\textbf{Quantitative comparison on light movement.} 
LightMover outperforms prior methods on the light-movement task in all metrics. * indicate one-step editing set up.}
\resizebox{0.95\columnwidth}{!}{
\small
\begin{tabular}{lccc}
\toprule
\textbf{Method} & \textbf{PSNR}$\uparrow$ & \textbf{DINO}$\uparrow$ & \textbf{CLIP}$\uparrow$ \\
\midrule
Qwen-Image~\cite{wu25_qwenimage} & 19.01 & 69.94 & 87.27 \\
Gemini-2.5-Flash-Image*~\cite{Gemini2.5Flash2025} & 18.18 & 52.06 & 82.21 \\
Gemini-2.5-Flash-Image~\cite{Gemini2.5Flash2025} & 19.59 & 72.46 & 89.72 \\
ObjectMover~\cite{yu25_objectmover} & 19.49 & 78.12 & 90.48 \\
\rowcolor{Cerulean!20}
\ours (ours) & 20.38 & \textbf{81.27} & \textbf{91.85} \\
\rowcolor{Cerulean!10}
\quad w/ adaptive pruning & \textbf{20.39} & 80.26 & 91.42 \\
\bottomrule
\end{tabular}
}
\label{tab:light_movement}
\end{table}

As shown in Table~\ref{tab:light_movement}, \ours significantly outperforms both LLM-powered editors and ObjectMover. Compared to text instruction-based models, \ours provides markedly higher precision in localizing and adjusting light positions, avoiding the ambiguous or global scene modifications typical of text-conditioned diffusion models. Compared to ObjectMover, our model exhibits superior \textbf{global illumination reasoning}. It can accurately capture reflection, shadow, and multi-light interactions to produce consistent and physically plausible light movement results.

\begin{figure}
    \centering
    \includegraphics[width=0.95\linewidth]{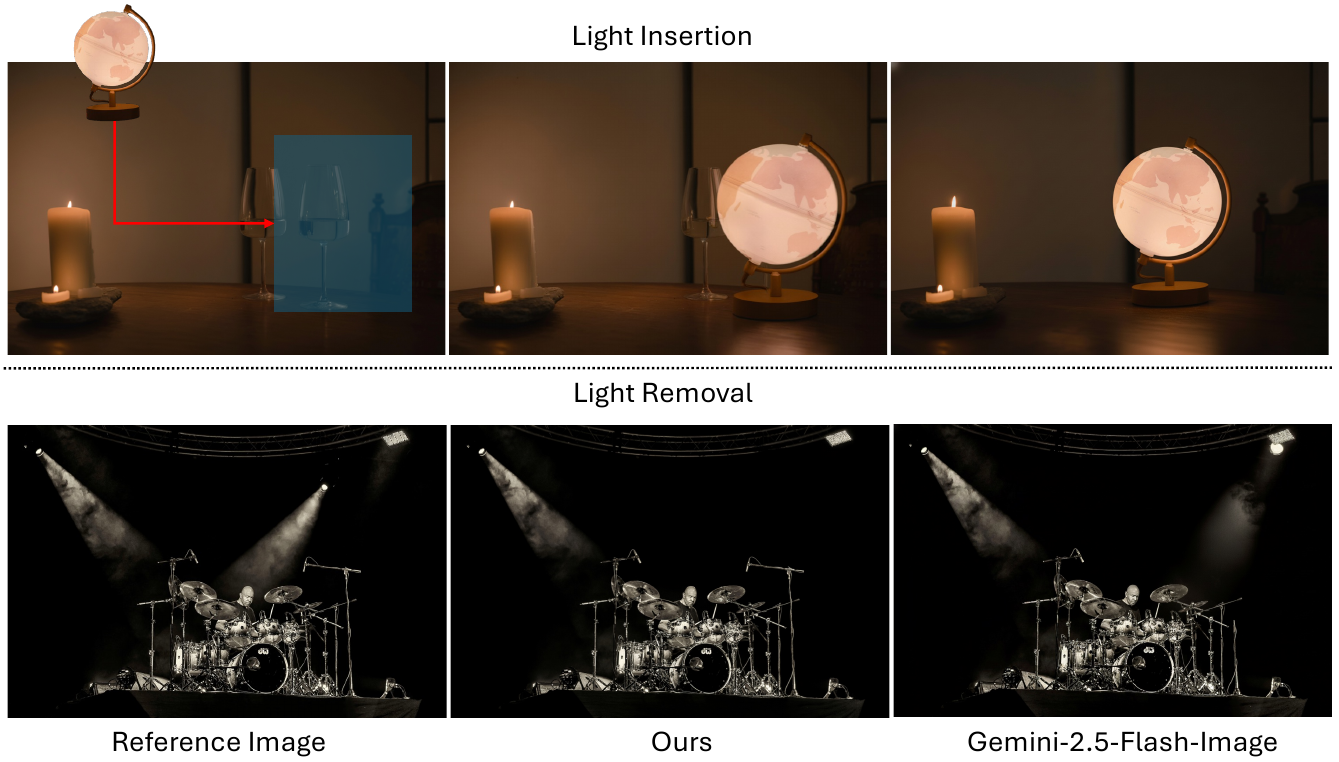}
    
    \caption{\textbf{Light insertion/removal results.} Compared to Nano-Banana~\cite{Gemini2.5Flash2025}, \ours produces more physically consistent edits, preserving background while correctly adding or removing illumination effects associated with the target light source.}
    \label{figs:insertion}
\end{figure}

Besides, we also present qualitative results on light insertion and removal in Figure~\ref{figs:insertion}. \ours better preserves the background scene and demonstrates a more coherent understanding of illumination composition, effectively removing all illumination-related effects.

\noindent\textbf{Light Movement with Illumination Control.}
Beyond spatial movement, we further evaluate \ours's ability to jointly control movement and illumination attributes using the synthetic \textit{LightMove-B} benchmark. The evaluation is divided into three folds:  
(1) \textit{Color change only}, where the light source is shifted among ten distinct target colors;  
(2) \textit{Intensity change only}, with ten target illumination levels; and  
(3) \textit{Joint movement with illumination change}, which combines spatial relocation with either color or intensity modulation.  
Each single-attribute split contains 100 paired examples, while the compositional split includes 150 pairs.

\begin{table*}[t]
\centering
\caption{\textbf{Quantitative comparison on light color change, intensity change, and their combination.}}
\resizebox{0.95\textwidth}{!}{
\small
\begin{tabular}{lcccccccccc}
\toprule
\multicolumn{1}{c}{\multirow{2}{*}{\textbf{Method}}} 
& \multicolumn{3}{c}{\textbf{Light Color Change}} 
& \multicolumn{3}{c}{\textbf{Light Intensity Change}} 
& \multicolumn{3}{c}{\textbf{Combine}} \\ 
\cmidrule(r){2-4}
\cmidrule(r){5-7}
\cmidrule(r){8-10}
& \textbf{PSNR}$\uparrow$ & \textbf{DINO}$\uparrow$ & \textbf{CLIP}$\uparrow$
& \textbf{PSNR}$\uparrow$ & \textbf{DINO}$\uparrow$ & \textbf{CLIP}$\uparrow$
& \textbf{PSNR}$\uparrow$ & \textbf{DINO}$\uparrow$ & \textbf{CLIP}$\uparrow$ \\
\midrule
Qwen-Image~\cite{wu25_qwenimage}
& 18.49 & 68.90 & 87.72 
& 20.41 & 71.05 & 88.96 
& 17.12 & 67.85 & 87.30 \\
Gemini-2.5-Flash-Image~\cite{Gemini2.5Flash2025}
& 22.14 & 81.72 & 88.65 
& 25.09 & 80.93 & 89.14 
& 18.42 & 76.21 & 88.22 \\
\rowcolor{Cerulean!20}
\ours (ours)
& 24.06 & \textbf{85.34} & 90.90
& \textbf{27.12} & 83.04 & \textbf{91.58}
& \textbf{19.97} & \textbf{79.86} & \textbf{91.02} \\
\rowcolor{Cerulean!10}
\quad w/ adaptive pruning
& \textbf{24.15} & 85.30 & \textbf{91.40}
& 27.01 & \textbf{83.11} & 91.21
& 19.88 & 78.92 & 90.66 \\
\bottomrule
\end{tabular}
}
\label{tab:light_other_tasks}
\end{table*}

As shown in Table~\ref{tab:light_other_tasks}, \ours consistently outperforms LLM-based baselines across all settings, demonstrating strong generalization to both individual and multiple light controls. Notably, our adaptive token pruning achieves comparable or superior accuracy in single-attribute tasks and yields substantial efficiency gains for multi-attribute sequences by reducing redundant tokens while preserving controllability.

\noindent\textbf{General Object Movement.}
\begin{table*}[t]
\centering
\caption{\textbf{Quantitative comparison on object movement, insertion, and removal.} Best results are in \textbf{bold} and second-best are \underline{underlined}.}
\resizebox{0.95\textwidth}{!}{
\definecolor{Gray}{gray}{0.94}
\begin{tabular}{lcccccccccccc}
\toprule
\multicolumn{1}{c}{\multirow{3}{*}{\textbf{Method}}} 
& \multicolumn{3}{c}{\textbf{Object Movement}} 
& \multicolumn{3}{c}{\textbf{Object Insertion}} 
& \multicolumn{3}{c}{\textbf{Object Removal}} \\ 
\cmidrule(r){2-4}
\cmidrule(r){5-7}
\cmidrule(r){8-10}
& \textbf{PSNR}$\uparrow$ & \textbf{DINO}$\uparrow$ & \textbf{CLIP}$\uparrow$
& \textbf{PSNR}$\uparrow$ & \textbf{DINO}$\uparrow$ & \textbf{CLIP}$\uparrow$
& \textbf{PSNR}$\uparrow$ & \textbf{DINO}$\uparrow$ & \textbf{CLIP}$\uparrow$ \\ 
\midrule
Drag-Anything~\cite{drag_anything} 
& 16.36 & 55.56 & 84.44 
& - & - & - 
& - & - & - \\
3DiT~\cite{3dit} 
& 19.72 & 45.30 & 81.69 
& - & - & - 
& 20.52 & 41.83 & 85.02 \\
Paint-by-Example~\cite{paint_by_example} 
& 20.83 & 55.46 & 85.23 
& 22.64 & 54.80 & 81.81 
& - & - & - \\
Anydoor~\cite{anydoor} 
& 21.86 & 69.32 & 88.95 
& {24.10} & {77.49} & {88.33}
& - & - & - \\
MagicFixup~\cite{magic_fixup} 
& {23.82} & {78.49} & {91.06}
& - & - & -
& - & - & - \\

ObjectMover~\cite{yu25_objectmover}
& {25.27} & {85.07} & \textbf{93.16}
& \underline{24.99} & {85.69} & \underline{91.45}
& \underline{28.90} & {83.94} & \textbf{94.84} \\

\rowcolor{Cerulean!20}
\ours (ours)
& \underline{25.73} & \textbf{88.59} & {91.86}
& \textbf{25.27} & \textbf{88.51} & {91.43}
& \underline{28.90} & \textbf{85.80} & \underline{93.94} \\
\rowcolor{Cerulean!10}\quad w/ adaptive pruning
& \textbf{25.74} & \underline{88.09} & \underline{92.74}
& 24.97 & \underline{88.01} & \textbf{92.21}
& \textbf{29.01} & \underline{84.74} & 93.85 \\
\bottomrule
\end{tabular}
}
\label{tab:merged_comparison}
\end{table*}

Beyond light source manipulation, we evaluate the generalization ability of \ours on the \textit{ObjMove-A} benchmark~\cite{yu25_objectmover}, following the evaluation protocol introduced in ObjectMover. This benchmark assesses general object \textit{movement}, \textit{removal}, and \textit{insertion}.  

We compare \ours on the core task of object movement against representative baselines, including \textit{MagicFixup}~\cite{magic_fixup}, \textit{3DiT}~\cite{3dit}, \textit{Drag-Anything}~\cite{drag_anything}, \textit{AnyDoor}~\cite{anydoor}, and \textit{Paint-by-Example}~\cite{paint_by_example}. For a fair comparison, insertion-based methods such as AnyDoor and Paint-by-Example are combined with an inpainting step to first remove the source object, while Drag-Anything generates the final frame of a short trajectory between source and target positions. Further implementation details are provided in the Supplementary.

As shown in Table~\ref{tab:merged_comparison}, \ours maintains strong performance on general object movement, consistently outperforming all baselines. This demonstrates that our unified formulation and joint training effectively generalizes beyond light-specific manipulation to broader spatial editing tasks.

\subsection{Ablation Studies}

\noindent\textbf{Effect of Training Data.}
\begin{table}[t]
\centering
\caption{\textbf{Ablation on training data composition for Light Movement.}  
We study the contribution of each augmentation: \textit{Light Aug.}, \textit{Color Change}, and \textit{Intensity Change}.}
\resizebox{0.95\columnwidth}{!}{
\definecolor{Gray}{gray}{0.94}
\definecolor{textGray}{gray}{0.6}

\begin{tabular}{cccccc}
\toprule
\multicolumn{3}{c}{\textbf{Training Data}} & 
\multicolumn{3}{c}{\textbf{Light Movement}} \\

\cmidrule(r){1-3}
\cmidrule(r){4-6}

\multicolumn{1}{c}{\textbf{Light Aug.}} & 
\multicolumn{1}{c}{\textbf{Color}} & 
\multicolumn{1}{c}{\textbf{Intensity}} &
\multicolumn{1}{c}{\textbf{PSNR}$\uparrow$} &
\multicolumn{1}{c}{\textbf{DINO}$\uparrow$} &
\multicolumn{1}{c}{\textbf{CLIP}$\uparrow$} \\

\midrule
\xmark & \xmark & \xmark &
19.88 & 79.93 & 91.06 \\
\cmark & \xmark & \xmark &
20.07 & 79.73 & \underline{91.62} \\
\cmark & \cmark & \xmark &
\underline{20.25} & \underline{81.02} & 91.02 \\
\cmark & \xmark & \cmark & 
20.11 & 80.73 & 90.66 \\
\cmark & \cmark & \cmark & 
\textbf{20.38} & \textbf{81.27} & \textbf{91.85} \\
\midrule
\end{tabular}
}
\label{tab:light_ablation}
\end{table}

We first study the impact of different training data configurations. We ablate variants trained with and without \textit{Light Aug.}, \textit{light color change}, and \textit{light intensity change} tasks. \textit{Light Aug.} denotes online illumination augmentation using physically disentangled rendering, where image pairs are dynamically rendered with varied light colors and intensities. As shown in Table~\ref{tab:light_ablation}, these auxiliary tasks are highly correlated and jointly enhance the model’s capability for precise light movement. Co-training with all tasks yields the best performance, as color and intensity variation help the model better disentangle the target light source’s contribution and learn the global light composition.

\noindent\textbf{Effect of Token Pruning Strategy.}
In this experiment, we investigate how different token pruning strategies affect performance when handling long control sequences. We evaluate three variants:  
(i) \emph{w/o frame-as-condition}: the light control attributes are not represented as image frames. Instead, two learned embeddings are introduced to encode color and intensity changes, which are concatenated to the token sequence.  
(ii) \emph{w/o object frame}: the object frame is removed from the control input to shorten the sequence length; and  
(iii) \emph{w/o adaptive downsampling}: all condition frames are uniformly downsampled to a fixed $8{\times}8$ resolution, removing the spatial adaptivity of our pruning scheme. As shown in Table~\ref{tab:auxiliary}, removing frame-based conditioning or spatially adaptive downsampling leads to a notable drop in controllability and visual fidelity. Our full model, equipped with the adaptive pruning strategy, achieves the best balance between efficiency and precision, demonstrating its importance for modeling complex multi-attribute illumination control.

\begin{table}[t]
\centering
\caption{\textbf{Ablation on Light Movement with color change.} We set \textbf{LightMover} as the baseline and study variants: (i) \emph{w/o frame as condition}, (ii) \emph{w/o object frame}, (iii) \emph{w/o adaptive downsampling}.}
\resizebox{0.95\linewidth}{!}{
\small
\begin{tabular}{l c c c}
\toprule
\textbf{Method} & \textbf{PSNR} $\uparrow$ & \textbf{DINO} $\uparrow$ & \textbf{CLIP} $\uparrow$ \\
\midrule
\rowcolor{Cerulean!20}
\textbf{LightMover (ours)} & \textbf{19.97} & \textbf{79.86} & \textbf{91.02} \\
\rowcolor{Cerulean!10}
\quad w/ adaptive pruning (ours) & \underline{19.88} & \underline{78.92} & \underline{90.66} \\
\midrule
\quad w/o frame as condition & 19.53 & 77.32 & 90.01 \\
\quad w/o object frame & 19.27 & 78.59 & 89.93 \\
\quad w/o adaptive downsample & 19.38 & 75.62 & 89.81 \\

\bottomrule
\end{tabular}
}
\label{tab:auxiliary}
\end{table}

\section{Conclusion}

This work demonstrates that physically consistent light manipulation can be achieved through a data-centric and learning-based approach. By constructing an automatic rendering pipeline for large-scale paired illumination data and leveraging a “2.5D” formulation that use video diffusion models for single-image light reasoning, \ours bridges the gap between fully 2D generative models and costly 3D physical rendering. We show that a learned model can approximate complex global light transport with far greater efficiency while maintaining photorealistic fidelity. We believe this paradigm opens new directions for scalable, physically aware image editing and for integrating learned illumination control into general 2D and 3D generative frameworks.

{
    \small
    \bibliographystyle{ieeenat_fullname}
    \bibliography{main}

\begin{thebibliography}{68}
\providecommand{\natexlab}[1]{#1}
\providecommand{\url}[1]{\texttt{#1}}
\expandafter\ifx\csname urlstyle\endcsname\relax
  \providecommand{\doi}[1]{doi: #1}\else
  \providecommand{\doi}{doi: \begingroup \urlstyle{rm}\Url}\fi

\bibitem[Aksoy et~al.(2018)Aksoy, Kim, Kellnhofer, Paris, Elgharib, Pollefeys, and Matusik]{aksoy2018flashambient}
Yagiz Aksoy, Changil Kim, Petr Kellnhofer, Sylvain Paris, Mohamed Elgharib, Marc Pollefeys, and Wojciech Matusik.
\newblock A dataset of flash and ambient illumination pairs from the crowd.
\newblock In \emph{Proceedings of the European Conference on Computer Vision (ECCV)}, pages 634--649, 2018.

\bibitem[Alzayer et~al.(2024)Alzayer, Xia, Zhang, Shechtman, Huang, and Gharbi]{magic_fixup}
Hadi Alzayer, Zhihao Xia, Xuaner Zhang, Eli Shechtman, Jia-Bin Huang, and Michael Gharbi.
\newblock Magic fixup: Streamlining photo editing by watching dynamic videos.
\newblock \emph{arXiv preprint arXiv:2403.13044}, 2024.

\bibitem[Avrahami et~al.(2022)Avrahami, Lischinski, and Fried]{avrahami2022blended}
Omri Avrahami, Dani Lischinski, and Ohad Fried.
\newblock Blended diffusion for text-driven editing of natural images.
\newblock In \emph{Proceedings of the IEEE/CVF conference on computer vision and pattern recognition}, pages 18208--18218, 2022.

\bibitem[Avrahami et~al.(2024)Avrahami, Gal, Chechik, Fried, Lischinski, Vahdat, and Nie]{avrahami2024diffuhaul}
Omri Avrahami, Rinon Gal, Gal Chechik, Ohad Fried, Dani Lischinski, Arash Vahdat, and Weili Nie.
\newblock Diffuhaul: A training-free method for object dragging in images.
\newblock \emph{arXiv preprint arXiv:2406.01594}, 2024.

\bibitem[Bhat et~al.(2024)Bhat, Mitra, and Wonka]{bhat2023loosecontrol}
Shariq~Farooq Bhat, Niloy~J. Mitra, and Peter Wonka.
\newblock Loosecontrol: Lifting controlnet for generalized depth conditioning.
\newblock In \emph{SIGGRAPH}, 2024.

\bibitem[Brooks et~al.(2023)Brooks, Holynski, and Efros]{brooks23_instructpix2pix}
Tim Brooks, Aleksander Holynski, and Alexei~A. Efros.
\newblock Instructpix2pix: Learning to follow image editing instructions.
\newblock In \emph{IEEE/CVF Conference on Computer Vision and Pattern Recognition (CVPR)}, 2023.

\bibitem[Careaga and Aksoy(2025)]{careaga2025physically}
Chris Careaga and Ya{\u{g}}{\i}z Aksoy.
\newblock Physically controllable relighting of photographs.
\newblock In \emph{Proceedings of the Special Interest Group on Computer Graphics and Interactive Techniques Conference Conference Papers}, pages 1--10, 2025.

\bibitem[Caron et~al.(2021)Caron, Touvron, Misra, J\'egou, Mairal, Bojanowski, and Joulin]{dino}
Mathilde Caron, Hugo Touvron, Ishan Misra, Herv\'e J\'egou, Julien Mairal, Piotr Bojanowski, and Armand Joulin.
\newblock Emerging properties in self-supervised vision transformers.
\newblock In \emph{Proceedings of the International Conference on Computer Vision (ICCV)}, 2021.

\bibitem[Chen et~al.(2024)Chen, Huang, Liu, Shen, Zhao, and Zhao]{anydoor}
Xi Chen, Lianghua Huang, Yu Liu, Yujun Shen, Deli Zhao, and Hengshuang Zhao.
\newblock Anydoor: Zero-shot object-level image customization.
\newblock In \emph{2024 IEEE/CVF Conference on Computer Vision and Pattern Recognition (CVPR)}, page 6593–6602. IEEE, 2024.

\bibitem[Community(2018)]{Blender2018}
Blender~Online Community.
\newblock Blender -- a 3d modelling and rendering package.
\newblock \url{http://www.blender.org}, 2018.
\newblock Blender Foundation, Stichting Blender Foundation, Amsterdam.

\bibitem[Couairon et~al.(2023)Couairon, Verbeek, Schwenk, and Cord]{couairon23_diffedit}
Guillaume Couairon, Jakob Verbeek, Holger Schwenk, and Matthieu Cord.
\newblock Diffedit: Diffusion-based semantic image editing with mask guidance.
\newblock In \emph{International Conference on Learning Representations (ICLR)}, 2023.

\bibitem[Debevec et~al.(2000)Debevec, Hawkins, Tchou, Duiker, Sarokin, and Sagar]{debevec2000acquiring}
Paul Debevec, Tim Hawkins, Chris Tchou, Haarm-Pieter Duiker, Westley Sarokin, and Mark Sagar.
\newblock Acquiring the reflectance field of a human face.
\newblock In \emph{Proceedings of the 27th annual conference on Computer graphics and interactive techniques}, pages 145--156, 2000.

\bibitem[Deitke et~al.(2023)Deitke, Liu, Wallingford, Ngo, Michel, Kusupati, Fan, Laforte, Voleti, Gadre, VanderBilt, Kembhavi, Vondrick, Gkioxari, Ehsani, Schmidt, and Farhadi]{objaverseXL}
Matt Deitke, Ruoshi Liu, Matthew Wallingford, Huong Ngo, Oscar Michel, Aditya Kusupati, Alan Fan, Christian Laforte, Vikram Voleti, Samir~Yitzhak Gadre, Eli VanderBilt, Aniruddha Kembhavi, Carl Vondrick, Georgia Gkioxari, Kiana Ehsani, Ludwig Schmidt, and Ali Farhadi.
\newblock Objaverse-xl: A universe of 10m+ 3d objects.
\newblock \emph{arXiv preprint arXiv:2307.05663}, 2023.

\bibitem[Dolhasz et~al.(2020)Dolhasz, Harvey, and Williams]{dolhasz2020learning}
Alan Dolhasz, Carlo Harvey, and Ian Williams.
\newblock Learning to observe: Approximating human perceptual thresholds for detection of suprathreshold image transformations.
\newblock In \emph{Proceedings of the IEEE/CVF Conference on Computer Vision and Pattern Recognition}, pages 4797--4807, 2020.

\bibitem[Geng and Owens(2024)]{geng2024motion}
Daniel Geng and Andrew Owens.
\newblock Motion guidance: Diffusion-based image editing with differentiable motion estimators.
\newblock \emph{International Conference on Learning Representations}, 2024.

\bibitem[{Google DeepMind}(2025)]{google25_gemini_flash_image}
{Google DeepMind}.
\newblock Introducing gemini 2.5 flash image: Image generation and editing with nano banana, 2025.
\newblock Google AI Studio / Gemini API blog.

\bibitem[Guo et~al.(2023)Guo, Wang, Yang, Huang, Wang, Pfister, and Wen]{guo2023shadowdiffusion}
Lanqing Guo, Chong Wang, Wenhan Yang, Siyu Huang, Yufei Wang, Hanspeter Pfister, and Bihan Wen.
\newblock Shadowdiffusion: When degradation prior meets diffusion model for shadow removal.
\newblock In \emph{Proceedings of the IEEE/CVF Conference on Computer Vision and Pattern Recognition}, pages 14049--14058, 2023.

\bibitem[Hui et~al.(2016)Hui, Sankaranarayanan, Sunkavalli, and Hadap]{hui2016whitebalance}
Zhuo Hui, Aswin~C Sankaranarayanan, Kalyan Sunkavalli, and Sunil Hadap.
\newblock White balance under mixed illumination using flash photography.
\newblock In \emph{2016 IEEE International Conference on Computational Photography (ICCP)}, pages 1--10. IEEE, 2016.

\bibitem[Hui et~al.(2018)Hui, Sunkavalli, Hadap, and Sankaranarayanan]{hui2017spectra}
Zhuo Hui, Kalyan Sunkavalli, Sunil Hadap, and Aswin~C Sankaranarayanan.
\newblock Illuminant spectra-based source separation using flash photography.
\newblock In \emph{Proceedings of the IEEE Conference on Computer Vision and Pattern Recognition}, pages 6209--6218, 2018.

\bibitem[Isola et~al.(2017)Isola, Zhu, Zhou, and Efros]{isola2017pix2pix}
Phillip Isola, Jun-Yan Zhu, Tinghui Zhou, and Alexei~A Efros.
\newblock Image-to-image translation with conditional adversarial networks.
\newblock In \emph{Proceedings of the IEEE conference on computer vision and pattern recognition}, pages 1125--1134, 2017.

\bibitem[Kingma(2014)]{adam}
Diederik~P Kingma.
\newblock Adam: A method for stochastic optimization.
\newblock \emph{arXiv preprint arXiv:1412.6980}, 2014.

\bibitem[Lensch et~al.(2003)Lensch, Kautz, Goesele, Heidrich, and Seidel]{lensch2003image}
Hendrik~PA Lensch, Jan Kautz, Michael Goesele, Wolfgang Heidrich, and Hans-Peter Seidel.
\newblock Image-based reconstruction of spatial appearance and geometric detail.
\newblock \emph{ACM Transactions on Graphics (TOG)}, 22\penalty0 (2):\penalty0 234--257, 2003.

\bibitem[Li et~al.(2020)Li, Shafiei, Ramamoorthi, Sunkavalli, and Chandraker]{li20_inverse_indoor}
Zhengqin Li, Mohammad Shafiei, Ravi Ramamoorthi, Kalyan Sunkavalli, and Manmohan Chandraker.
\newblock Inverse rendering for complex indoor scenes: Shape, spatially-varying lighting and {SVBRDF} from a single image.
\newblock In \emph{IEEE/CVF Conference on Computer Vision and Pattern Recognition (CVPR)}, 2020.

\bibitem[Li et~al.(2022)Li, Shi, Bi, Zhu, Sunkavalli, Ha{\v{s}}an, Xu, Ramamoorthi, and Chandraker]{li22_indoor_light}
Zhengqin Li, Jia Shi, Sai Bi, Rui Zhu, Kalyan Sunkavalli, Milo{\v{s}} Ha{\v{s}}an, Zexiang Xu, Ravi Ramamoorthi, and Manmohan Chandraker.
\newblock Physically-based editing of indoor scene lighting from a single image.
\newblock In \emph{European Conference on Computer Vision (ECCV)}, 2022.

\bibitem[Liang et~al.(2025)Liang, Gojcic, Ling, Munkberg, Hasselgren, Lin, Gao, Keller, Vijaykumar, Fidler, et~al.]{liang2025diffusion}
Ruofan Liang, Zan Gojcic, Huan Ling, Jacob Munkberg, Jon Hasselgren, Chih-Hao Lin, Jun Gao, Alexander Keller, Nandita Vijaykumar, Sanja Fidler, et~al.
\newblock Diffusion renderer: Neural inverse and forward rendering with video diffusion models.
\newblock In \emph{Proceedings of the Computer Vision and Pattern Recognition Conference}, pages 26069--26080, 2025.

\bibitem[Lipman et~al.(2022)Lipman, Chen, Ben-Hamu, Nickel, and Le]{flow_matching}
Yaron Lipman, Ricky~TQ Chen, Heli Ben-Hamu, Maximilian Nickel, and Matt Le.
\newblock Flow matching for generative modeling.
\newblock \emph{arXiv preprint arXiv:2210.02747}, 2022.

\bibitem[Liu et~al.(2023)Liu, Zhang, Ma, Peng, et~al.]{instaflow}
Xingchao Liu, Xiwen Zhang, Jianzhu Ma, Jian Peng, et~al.
\newblock Instaflow: One step is enough for high-quality diffusion-based text-to-image generation.
\newblock In \emph{The Twelfth International Conference on Learning Representations}, 2023.

\bibitem[Lu et~al.(2023)Lu, Liu, and Kong]{lu2023tficon}
Shilin Lu, Yanzhu Liu, and Adams Wai-Kin Kong.
\newblock Tf-icon: Diffusion-based training-free cross-domain image composition.
\newblock In \emph{Proceedings of the IEEE/CVF International Conference on Computer Vision}, 2023.

\bibitem[Lugmayr et~al.(2022)Lugmayr, Danelljan, Romero, Yu, Timofte, and Van~Gool]{lugmayr2022repaint}
Andreas Lugmayr, Martin Danelljan, Andres Romero, Fisher Yu, Radu Timofte, and Luc Van~Gool.
\newblock Repaint: Inpainting using denoising diffusion probabilistic models.
\newblock In \emph{Proceedings of the IEEE/CVF conference on computer vision and pattern recognition}, pages 11461--11471, 2022.

\bibitem[Magar et~al.(2025)Magar, Hertz, Tabellion, Pritch, Rav-Acha, Shamir, and Hoshen]{magar2025lightlab}
Nadav Magar, Amir Hertz, Eric Tabellion, Yael Pritch, Alex Rav-Acha, Ariel Shamir, and Yedid Hoshen.
\newblock Lightlab: Controlling light sources in images with diffusion models.
\newblock In \emph{Proceedings of the Special Interest Group on Computer Graphics and Interactive Techniques Conference Conference Papers}, pages 1--11, 2025.

\bibitem[Maralan et~al.(2023)Maralan, Careaga, and Aksoy]{maralan2023flashintrinsics}
Sepideh~Sarajian Maralan, Chris Careaga, and Yagiz Aksoy.
\newblock Computational flash photography through intrinsics.
\newblock In \emph{Proceedings of the IEEE/CVF Conference on Computer Vision and Pattern Recognition}, pages 16654--16662, 2023.

\bibitem[Meng et~al.(2022)Meng, He, Song, Song, Wu, Zhu, and Ermon]{meng2021sdedit}
Chenlin Meng, Yutong He, Yang Song, Jiaming Song, Jiajun Wu, Jun-Yan Zhu, and Stefano Ermon.
\newblock Sdedit: Guided image synthesis and editing with stochastic differential equations.
\newblock \emph{ICLR}, 2022.

\bibitem[Michel et~al.(2024)Michel, Bhattad, VanderBilt, Krishna, Kembhavi, and Gupta]{3dit}
Oscar Michel, Anand Bhattad, Eli VanderBilt, Ranjay Krishna, Aniruddha Kembhavi, and Tanmay Gupta.
\newblock Object 3dit: Language-guided 3d-aware image editing.
\newblock \emph{Advances in Neural Information Processing Systems}, 36, 2024.

\bibitem[Mou et~al.(2024{\natexlab{a}})Mou, Wang, Song, Shan, and Zhang]{mou2023dragondiffusion}
Chong Mou, Xintao Wang, Jiechong Song, Ying Shan, and Jian Zhang.
\newblock Dragondiffusion: Enabling drag-style manipulation on diffusion models.
\newblock \emph{ICLR}, 2024{\natexlab{a}}.

\bibitem[Mou et~al.(2024{\natexlab{b}})Mou, Wang, Song, Shan, and Zhang]{mou2024diffeditor}
Chong Mou, Xintao Wang, Jiechong Song, Ying Shan, and Jian Zhang.
\newblock Diffeditor: Boosting accuracy and flexibility on diffusion-based image editing.
\newblock In \emph{Proceedings of the IEEE/CVF Conference on Computer Vision and Pattern Recognition}, pages 8488--8497, 2024{\natexlab{b}}.

\bibitem[Murmann et~al.(2016)Murmann, Davis, Kautz, and Durand]{murmann2016bounceflash}
Lukas Murmann, Abe Davis, Jan Kautz, and Fr{\'e}do Durand.
\newblock Computational bounce flash for indoor portraits.
\newblock \emph{ACM Transactions on Graphics (TOG)}, 35\penalty0 (6):\penalty0 1--9, 2016.

\bibitem[Murmann et~al.(2019)Murmann, Gharbi, Aittala, and Durand]{murmann2019multiillumination}
Lukas Murmann, Michael Gharbi, Miika Aittala, and Fredo Durand.
\newblock A dataset of multi-illumination images in the wild.
\newblock In \emph{Proceedings of the IEEE/CVF International Conference on Computer Vision}, pages 4080--4089, 2019.

\bibitem[{OpenAI}(2023)]{sora}
{OpenAI}.
\newblock Video generation models as world simulators.
\newblock \url{https://openai.com/research/video-generation-models-as-world-simulators}, 2023.

\bibitem[Pandey et~al.(2024)Pandey, Guerrero, Gadelha, Hold-Geoffroy, Singh, and Mitra]{diff_handle}
Karran Pandey, Paul Guerrero, Matheus Gadelha, Yannick Hold-Geoffroy, Karan Singh, and Niloy~J. Mitra.
\newblock Diffusion handles enabling 3d edits for diffusion models by lifting activations to 3d.
\newblock In \emph{2024 IEEE/CVF Conference on Computer Vision and Pattern Recognition (CVPR)}, page 7695–7704. IEEE, 2024.

\bibitem[Peebles and Xie(2023)]{dit}
William Peebles and Saining Xie.
\newblock Scalable diffusion models with transformers.
\newblock In \emph{Proceedings of the IEEE/CVF International Conference on Computer Vision}, pages 4195--4205, 2023.

\bibitem[Peng et~al.(2023)Peng, Quesnelle, Fan, and Shippole]{peng2023yarn}
Bowen Peng, Jeffrey Quesnelle, Honglu Fan, and Enrico Shippole.
\newblock Yarn: Efficient context window extension of large language models.
\newblock \emph{arXiv preprint arXiv:2309.00071}, 2023.

\bibitem[Pexels(2024)]{Pexels}
Pexels.
\newblock Pexels - free high-quality images.
\newblock \url{https://www.pexels.com/}, 2024.

\bibitem[{Qwen Team}(2025)]{qwen25_image_edit}
{Qwen Team}.
\newblock Qwen-image-edit: Open-source image editing model.
\newblock \url{https://qwenimages.com/blog/qwen-image-edit-release}, 2025.
\newblock Model and blog post.

\bibitem[Radford et~al.(2021)Radford, Kim, Hallacy, Ramesh, Goh, Agarwal, Sastry, Askell, Mishkin, Clark, et~al.]{clip}
Alec Radford, Jong~Wook Kim, Chris Hallacy, Aditya Ramesh, Gabriel Goh, Sandhini Agarwal, Girish Sastry, Amanda Askell, Pamela Mishkin, Jack Clark, et~al.
\newblock Learning transferable visual models from natural language supervision.
\newblock In \emph{International conference on machine learning}, pages 8748--8763. PMLR, 2021.

\bibitem[Research(2025)]{Gemini2.5Flash2025}
Google Research.
\newblock Gemini 2.5 flash — image model.
\newblock \url{https://aistudio.google.com/models/gemini-2-5-flash-image}, 2025.
\newblock AI Studio, Google Research.

\bibitem[Ronneberger et~al.(2015)Ronneberger, Fischer, and Brox]{ronneberger2015unet}
Olaf Ronneberger, Philipp Fischer, and Thomas Brox.
\newblock U-net: Convolutional networks for biomedical image segmentation.
\newblock In \emph{International Conference on Medical image computing and computer-assisted intervention}, pages 234--241. Springer, 2015.

\bibitem[Saharia et~al.(2022)Saharia, Chan, Chang, Lee, Ho, Salimans, Fleet, and Norouzi]{saharia2022palette}
Chitwan Saharia, William Chan, Huiwen Chang, Chris Lee, Jonathan Ho, Tim Salimans, David Fleet, and Mohammad Norouzi.
\newblock Palette: Image-to-image diffusion models.
\newblock In \emph{ACM SIGGRAPH 2022 conference proceedings}, pages 1--10, 2022.

\bibitem[Sengupta et~al.(2019)Sengupta, Gu, Kim, Liu, Jacobs, and Kautz]{sengupta19_nir}
Soumyadip Sengupta, Jinwei Gu, Kihwan Kim, Guilin Liu, David~W. Jacobs, and Jan Kautz.
\newblock Neural inverse rendering of an indoor scene from a single image.
\newblock In \emph{IEEE/CVF International Conference on Computer Vision (ICCV)}, 2019.

\bibitem[Shi et~al.(2024)Shi, Xue, Liew, Pan, Yan, Zhang, Tan, and Bai]{shi2024dragdiffusion}
Yujun Shi, Chuhui Xue, Jun~Hao Liew, Jiachun Pan, Hanshu Yan, Wenqing Zhang, Vincent~YF Tan, and Song Bai.
\newblock Dragdiffusion: Harnessing diffusion models for interactive point-based image editing.
\newblock In \emph{Proceedings of the IEEE/CVF Conference on Computer Vision and Pattern Recognition}, pages 8839--8849, 2024.

\bibitem[Song et~al.(2023)Song, Zhang, Lin, Cohen, Price, Zhang, Kim, and Aliaga]{objectstitch}
Yizhi Song, Zhifei Zhang, Zhe Lin, Scott Cohen, Brian Price, Jianming Zhang, Soo~Ye Kim, and Daniel Aliaga.
\newblock Objectstitch: Object compositing with diffusion model.
\newblock In \emph{2023 IEEE/CVF Conference on Computer Vision and Pattern Recognition (CVPR)}. IEEE, 2023.

\bibitem[Song et~al.(2024)Song, Zhang, Lin, Cohen, Price, Zhang, Kim, Zhang, Xiong, and Aliaga]{imprint}
Yizhi Song, Zhifei Zhang, Zhe Lin, Scott Cohen, Brian Price, Jianming Zhang, Soo~Ye Kim, He Zhang, Wei Xiong, and Daniel Aliaga.
\newblock Imprint: Generative object compositing by learning identity-preserving representation.
\newblock In \emph{2024 IEEE/CVF Conference on Computer Vision and Pattern Recognition (CVPR)}, page 8048–8058. IEEE, 2024.

\bibitem[Su et~al.(2024)Su, Ahmed, Lu, Pan, Bo, and Liu]{su2024roformer}
Jianlin Su, Murtadha Ahmed, Yu Lu, Shengfeng Pan, Wen Bo, and Yunfeng Liu.
\newblock Roformer: Enhanced transformer with rotary position embedding.
\newblock \emph{Neurocomputing}, 568:\penalty0 127063, 2024.

\bibitem[Suvorov et~al.(2021)Suvorov, Logacheva, Mashikhin, Remizova, Ashukha, Silvestrov, Kong, Goka, Park, and Lempitsky]{suvorov2021resolution}
Roman Suvorov, Elizaveta Logacheva, Anton Mashikhin, Anastasia Remizova, Arsenii Ashukha, Aleksei Silvestrov, Naejin Kong, Harshith Goka, Kiwoong Park, and Victor Lempitsky.
\newblock Resolution-robust large mask inpainting with fourier convolutions.
\newblock \emph{arXiv preprint arXiv:2109.07161}, 2021.

\bibitem[Tarrés et~al.(2024)Tarrés, Lin, Zhang, Zhang, Song, Ruta, Gilbert, Collomosse, and Kim]{think_outside}
Gemma~Canet Tarrés, Zhe Lin, Zhifei Zhang, Jianming Zhang, Yizhi Song, Dan Ruta, Andrew Gilbert, John Collomosse, and Soo~Ye Kim.
\newblock Thinking outside the bbox: Unconstrained generative object compositing, 2024.

\bibitem[Vaswani(2017)]{transformer}
A Vaswani.
\newblock Attention is all you need.
\newblock \emph{Advances in Neural Information Processing Systems}, 2017.

\bibitem[Wang et~al.(2025)Wang, Tran, Cui, Thomson, Dahl, Arjomand~Bigdeli, Frisvad, and Chandraker]{wang25_materialist}
Lezhong Wang, Duc~Minh Tran, Ruiqi Cui, T.~G. Thomson, Anders~Bjorholm Dahl, Siavash Arjomand~Bigdeli, Jeppe~Revall Frisvad, and Manmohan Chandraker.
\newblock Materialist: Physically based editing using single-image inverse rendering, 2025.

\bibitem[Wang et~al.(2020)Wang, Hu, Wang, Heng, and Fu]{wang2020instance}
Tianyu Wang, Xiaowei Hu, Qiong Wang, Pheng-Ann Heng, and Chi-Wing Fu.
\newblock Instance shadow detection.
\newblock In \emph{Proceedings of the IEEE/CVF Conference on Computer Vision and Pattern Recognition}, pages 1880--1889, 2020.

\bibitem[Winter et~al.(2024)Winter, Cohen, Fruchter, Pritch, Rav-Acha, and Hoshen]{objectdrop}
Daniel Winter, Matan Cohen, Shlomi Fruchter, Yael Pritch, Alex Rav-Acha, and Yedid Hoshen.
\newblock Objectdrop: Bootstrapping counterfactuals for photorealistic object removal and insertion, 2024.

\bibitem[Wu et~al.(2025{\natexlab{a}})Wu, Li, Zhou, Lin, Gao, Yan, Yin, Bai, Xu, Chen, et~al.]{wu25_qwenimage}
Chenfei Wu, Jiahao Li, Jingren Zhou, Junyang Lin, Kaiyuan Gao, Kun Yan, Sheng-ming Yin, Shuai Bai, Xiao Xu, Yilei Chen, et~al.
\newblock Qwen-image technical report.
\newblock \emph{arXiv preprint arXiv:2508.02324}, 2025{\natexlab{a}}.

\bibitem[Wu et~al.(2025{\natexlab{b}})Wu, Basu, Br{\"o}dermann, Van~Gool, and Sakaridis]{wu25_pbrnerf}
Sean Wu, Shamik Basu, Tim Br{\"o}dermann, Luc Van~Gool, and Christos Sakaridis.
\newblock {PBR}-{NeRF}: Inverse rendering with physics-based neural fields.
\newblock In \emph{IEEE/CVF Conference on Computer Vision and Pattern Recognition (CVPR)}, 2025{\natexlab{b}}.

\bibitem[Wu et~al.(2025{\natexlab{c}})Wu, Li, Gu, Zhao, He, Zhang, Shou, Li, Gao, and Zhang]{drag_anything}
Weijia Wu, Zhuang Li, Yuchao Gu, Rui Zhao, Yefei He, David~Junhao Zhang, Mike~Zheng Shou, Yan Li, Tingting Gao, and Di Zhang.
\newblock Draganything: Motion control for anything using entity representation.
\newblock In \emph{European Conference on Computer Vision}, pages 331--348. Springer, 2025{\natexlab{c}}.

\bibitem[Xing et~al.(2025)Xing, Groh, Karaoglu, Gevers, and Bhattad]{xing2025luminet}
Xiaoyan Xing, Konrad Groh, Sezer Karaoglu, Theo Gevers, and Anand Bhattad.
\newblock Luminet: Latent intrinsics meets diffusion models for indoor scene relighting.
\newblock In \emph{Proceedings of the Computer Vision and Pattern Recognition Conference}, pages 442--452, 2025.

\bibitem[Yang et~al.(2023)Yang, Gu, Zhang, Zhang, Chen, Sun, Chen, and Wen]{paint_by_example}
Binxin Yang, Shuyang Gu, Bo Zhang, Ting Zhang, Xuejin Chen, Xiaoyan Sun, Dong Chen, and Fang Wen.
\newblock Paint by example: Exemplar-based image editing with diffusion models.
\newblock In \emph{2023 IEEE/CVF Conference on Computer Vision and Pattern Recognition (CVPR)}. IEEE, 2023.

\bibitem[Yenphraphai et~al.(2024)Yenphraphai, Pan, Liu, Panozzo, and Xie]{yenphraphai2024image}
Jiraphon Yenphraphai, Xichen Pan, Sainan Liu, Daniele Panozzo, and Saining Xie.
\newblock Image sculpting: Precise object editing with 3d geometry control.
\newblock In \emph{Proceedings of the IEEE/CVF Conference on Computer Vision and Pattern Recognition}, pages 4241--4251, 2024.

\bibitem[Yu et~al.(2025)Yu, Wang, Kim, Guerrero, Chen, Liu, Lin, and Qi]{yu25_objectmover}
Xin Yu, Tianyu Wang, Soo~Ye Kim, Paul Guerrero, Xi Chen, Qing Liu, Zhe Lin, and Xiaojuan Qi.
\newblock {ObjectMover}: Generative object movement with video prior.
\newblock In \emph{Proceedings of the {IEEE/CVF} Conference on Computer Vision and Pattern Recognition (CVPR)}, 2025.

\bibitem[Zeng et~al.(2024)Zeng, Deschaintre, Georgiev, Hold-Geoffroy, Hu, Luan, Yan, and Ha\v{s}an]{zeng2024rgb}
Zheng Zeng, Valentin Deschaintre, Iliyan Georgiev, Yannick Hold-Geoffroy, Yiwei Hu, Fujun Luan, Ling-Qi Yan, and Milo\v{s} Ha\v{s}an.
\newblock Rgb-x: Image decomposition and synthesis using material- and lighting-aware diffusion models.
\newblock In \emph{ACM SIGGRAPH 2024 Conference Papers}, New York, NY, USA, 2024. Association for Computing Machinery.

\bibitem[Zhang et~al.(2023)Zhang, Duan, Lan, Hong, Zhu, Wang, and Niu]{zhang2023controlcom}
Bo Zhang, Yuxuan Duan, Jun Lan, Yan Hong, Huijia Zhu, Weiqiang Wang, and Li Niu.
\newblock Controlcom: Controllable image composition using diffusion model.
\newblock \emph{arXiv preprint arXiv:2308.10040}, 2023.

\bibitem[Zhang et~al.(2025)Zhang, Rao, and Agrawala]{zhang2025scaling}
Lvmin Zhang, Anyi Rao, and Maneesh Agrawala.
\newblock Scaling in-the-wild training for diffusion-based illumination harmonization and editing by imposing consistent light transport.
\newblock In \emph{The Thirteenth International Conference on Learning Representations}, 2025.

\end{thebibliography}
}

\clearpage
\setcounter{page}{1}
\maketitlesupplementary

\appendix

\section{Implementation Details}
\label{sec:impl_details}

\noindent\textbf{Training.}
The model is trained on 16 NVIDIA A100 GPUs for 15{,}000 iterations with a batch size of 32. We use the AdamW optimizer~\cite{adam} with a weight decay of $0.01$ and an initial learning rate of $1 \times 10^{-4}$. Exponential Moving Average (EMA) with a decay rate of $0.99$ is applied after the first 1{,}000 iterations to stabilize training.

\noindent\textbf{Data Sampling.}
Training samples are drawn at mixed resolutions of $512 \times 512$ and $1024 \times 1024$ pixels with a 1:1 ratio. The dataset combines synthetic and real data at a 10:1 ratio. Synthetic tasks are distributed across seven categories: (1)~light movement, (2)~object movement, (3)~light color change, (4)~light intensity change, (5)~joint movement and color/intensity change, (6)~light removal, and (7)~light insertion, following a ratio of 6:3:3:3:1:1, respectively.

\noindent\textbf{Augmentation.}
To improve robustness to spatial variation, we apply box augmentation by maintaining the bounding-box center while randomly scaling its dimensions within $[0.8, 1.2]$. We additionally apply light-illumination augmentation via the Physically Disentangled Rendering procedure (Sec.~3.2 of the main paper), which dynamically varies ambient and direct light components during training.

\section{Additional Comparisons and Ablations}
\label{sec:additional_results}

We present additional comparisons and ablation visualizations that complement the quantitative results in the main paper.

\subsection{Comparison with Ground Truth}

\begin{figure*}[t]
    \centering
    \includegraphics[width=\linewidth]{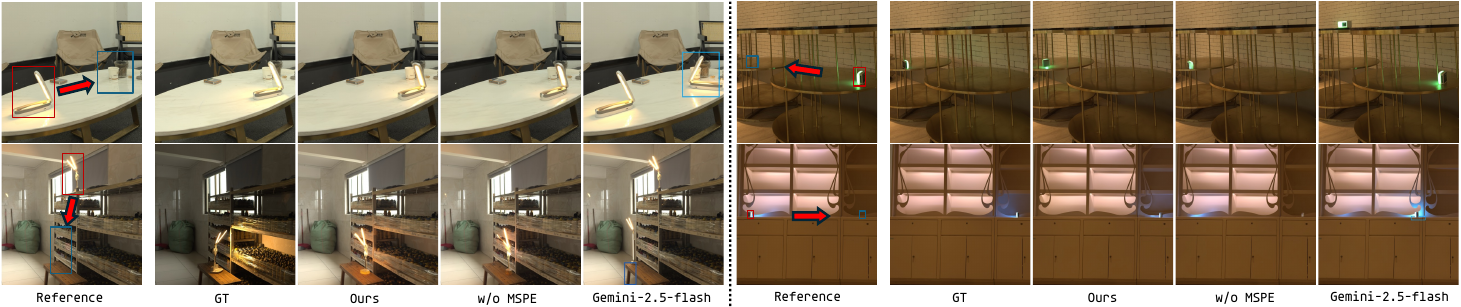}
    \caption{\textbf{Comparison with ground-truth images.} We show side-by-side comparisons between \ours predictions and ground-truth photographs captured by the same camera under the same ambient conditions but with the light source physically relocated.}
    \label{figs:comp_gt}
\end{figure*}

To validate the physical plausibility of \ours, we directly compare our predictions against real ground-truth photographs from the \textit{LightMove-A} benchmark, where each ground truth is captured by the same camera under identical ambient conditions with the light source physically relocated. As shown in Figure~\ref{figs:comp_gt}, \ours achieves precise positional control for local lighting changes, accurately reproducing shadows, highlights, and surface shading patterns that closely match the ground-truth images. Notably, these examples are selected without cherry-picking from challenging in-the-wild scenes, demonstrating that our model generalizes robustly to diverse real-world environments without introducing visual hallucinations or physically inconsistent artifacts.

\subsection{Visual Ablation of Training Data Composition}

\begin{figure*}[t]
    \centering
    \includegraphics[width=\linewidth]{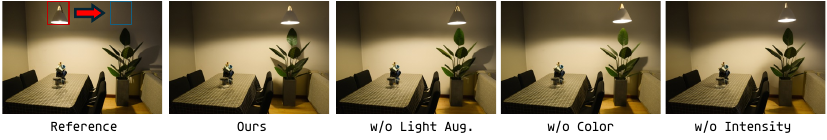}
    \caption{\textbf{Qualitative results for Table~4 (training data ablation).} With the proposed synthetic rendering data pipeline (Light Aug.), the model generalizes to complex shadow handling (e.g., leaf shading). Incorporating color and intensity variations further improves the modeling of light contributions and their combinations.}
    \label{figs:vis_tab4}
\end{figure*}

Figure~\ref{figs:vis_tab4} provides qualitative ablation results corresponding to the quantitative analysis in Table~4 of the main paper. These visual comparisons illustrate the effectiveness of the proposed data augmentation strategy and the synergistic effects of training data composition. Specifically, training with the physically disentangled rendering augmentation (\textit{Light Aug.}) enables the model to generalize to complex shadow interactions, such as intricate leaf shading patterns, that are absent in the unaugmented data. Furthermore, incorporating light color and intensity variation tasks during training improves the model's understanding of how individual light sources contribute to the overall scene illumination. The full model, trained with all proposed data components, produces the most physically coherent results by effectively disentangling and recombining illumination effects.

\subsection{Comparison with the Learnable Embedding Variant}

\begin{figure*}[h]
    \centering
    \includegraphics[width=\linewidth]{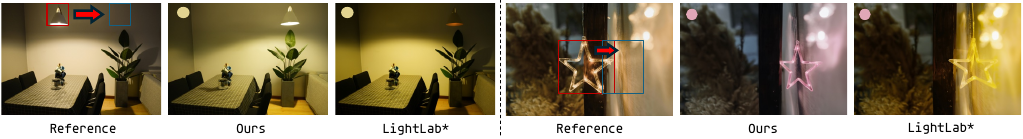}
    \caption{\textbf{Comparison with the learnable embedding variant.} We evaluate the move-with-color-change task, where the colored dot in the top-left indicates the target color. The LightLab variant with learnable embeddings fails to correctly encode color conditions when movement and color changes are applied together.}
    \label{figs:lightlab}
\end{figure*}

We compare our frame-based conditioning design against a variant that uses separate learnable embeddings for different control signals (Table~5, line~3 in the main paper). As shown in Figure~\ref{figs:lightlab}, we evaluate the joint move-with-color-change task, where the colored dot in the top-left corner of each result indicates the target light color. The variant with condition-specific learnable embeddings fails to correctly recognize color conditions when movement and color changes are applied simultaneously. This is because learnable embeddings are less flexible and do not generalize well to multiple simultaneous controls: they tend to conflate or ignore individual signals when multiple conditions are introduced jointly. In contrast, our frame-based conditioning naturally integrates with the video diffusion backbone and supports joint control over movement, color, and intensity by encoding each condition as an explicit image frame, enabling the model to maintain distinct and accurate representations for each control signal.

\section{Prompt Details for Nano-Banana Light-Movement Baselines}
\label{sec:nanobanana_prompts}

As described in the main paper (Sec.~4), we evaluate two representative LLM-powered text-to-image editing setups for Gemini-2.5-Flash-Image:  
(1) a \emph{one-step} editing pipeline, and  
(2) a \emph{two-step} remove--insert pipeline.  
For completeness, we provide the exact prompt templates used in both settings, following the Gemini-2.5-Flash-Image's generation results.

\subsection{One-Step Editing}
In the one-step setup, both the original light-source location and the target location are marked in the input image using a \textcolor{red}{red} bounding box and a \textcolor{blue}{blue} bounding box, respectively.  
The model receives a single prompt instructing it to relocate the light source from the red region to the blue region and produce the corresponding physically consistent relighting.  
We include three variants: \emph{concise}, \emph{basic}, and \emph{precise} in our preliminary experiments. Detail prompts are shown below:

\begin{figure}[h]
    \centering
    \begin{tcolorbox}[
    colback=NavyBlue!5!white,
    colframe=NavyBlue!5!white,
    arc=2mm,
    boxrule=0pt,
    left=2mm, right=2mm, top=2mm, bottom=2mm,
    fontupper=\scriptsize
]
\textcolor{pink}{\{reference\_image\}}

\textcolor{RoyalBlue}{Move the light source from the RED box to the BLUE box, then remove all bounding boxes.}
    \end{tcolorbox}
    \caption{Concise prompt for the one-step editing.}
    \label{fig:onestep_basic}
\end{figure}

\begin{figure}[h]
    \centering
    \begin{tcolorbox}[
    colback=NavyBlue!5!white,
    colframe=NavyBlue!5!white,
    arc=2mm,
    boxrule=0pt,
    left=2mm, right=2mm, top=2mm, bottom=2mm,
    fontupper=\scriptsize
]
\textcolor{pink}{\{reference\_image\}}

\textcolor{RoyalBlue}{You are an expert image editor specialized in realistic light-source manipulation and relighting. Your task is to move a light source from one location to another while maintaining physical realism.}

Key considerations:

1. Ensure physically accurate lighting behavior \\
2. Shadows and highlights must update based on the new light position \\
3. Preserve global scene consistency \\
4. Maintain light intensity and color temperature unless specified \\
5. Keep material properties (reflectance, specularity) unchanged
    \end{tcolorbox}
    \caption{\textbf{Basic prompt variant for the one-step editing.}  
    This prompt emphasizes physical correctness and globally consistent relighting.}
    \label{fig:onestep_detailed}
\end{figure}

\begin{figure}[h]
    \centering
    \begin{tcolorbox}[
    colback=NavyBlue!5!white,
    colframe=NavyBlue!5!white,
    arc=2mm,
    boxrule=0pt,
    left=2mm, right=2mm, top=2mm, bottom=2mm,
    fontupper=\scriptsize
]
\textcolor{pink}{\{reference\_image\}}

\textcolor{RoyalBlue}{Reference Image Analysis:}

- RED BOX: Current position of the light source

- BLUE BOX: Desired target position for the light source\\

\textcolor{PineGreen}{Task: Light Source Relocation with Realistic Relighting}

\textcolor{RoyalBlue}{Step-by-step requirements:}

1. Object Movement:

   - Move the light source object from the red box to the blue box position
   
   - Maintain the object's orientation and appearance\\

2. Lighting Updates:

   - Calculate new shadow directions based on the target light position
   
   - Update highlights on reflective surfaces
   
   - Adjust ambient lighting in the scene
   
   - Ensure shadows are cast from the new light position\\

3. Physical Realism:

   - Maintain consistent light intensity
   
   - Keep natural shadow softness based on light source size
   
   - Preserve material properties of all objects
   
   - Ensure proper light falloff with distance\\

4. Scene Consistency:

   - Keep the background unchanged except for lighting
   
   - Maintain color temperature
   
   - Preserve image resolution and quality\\

Generate the photorealistic result with the light source successfully relocated.
    \end{tcolorbox}
    \caption{\textbf{``Precise'' step-by-step prompt variant for the one-step editing.}  
    This version enforces explicit reasoning over shadow geometry, reflectance behavior, and scene consistency.}
    \label{fig:onestep_precise}
\end{figure}

\begin{figure*}
    \centering
    \includegraphics[width=0.9\linewidth]{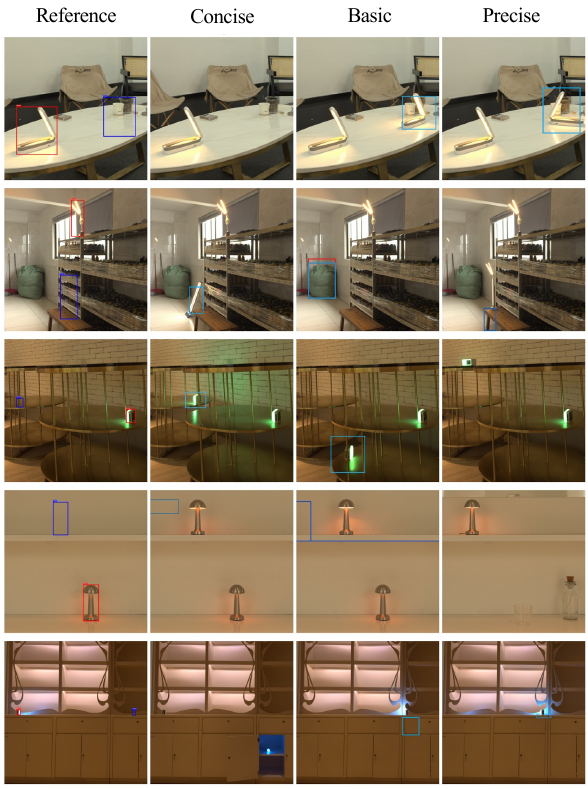}
    
    \caption{\textbf{Light movement results for Nano-Banana~\cite{Gemini2.5Flash2025} one-step editing.} We show the input reference image and the results of using different prompts.}
    \label{figs:onestep}
\end{figure*}

We show the generated results in Figure~\ref{figs:onestep}.

\subsection{Two-Step Editing}

\begin{figure*}[t]
    \centering
    \includegraphics[width=0.95\linewidth]{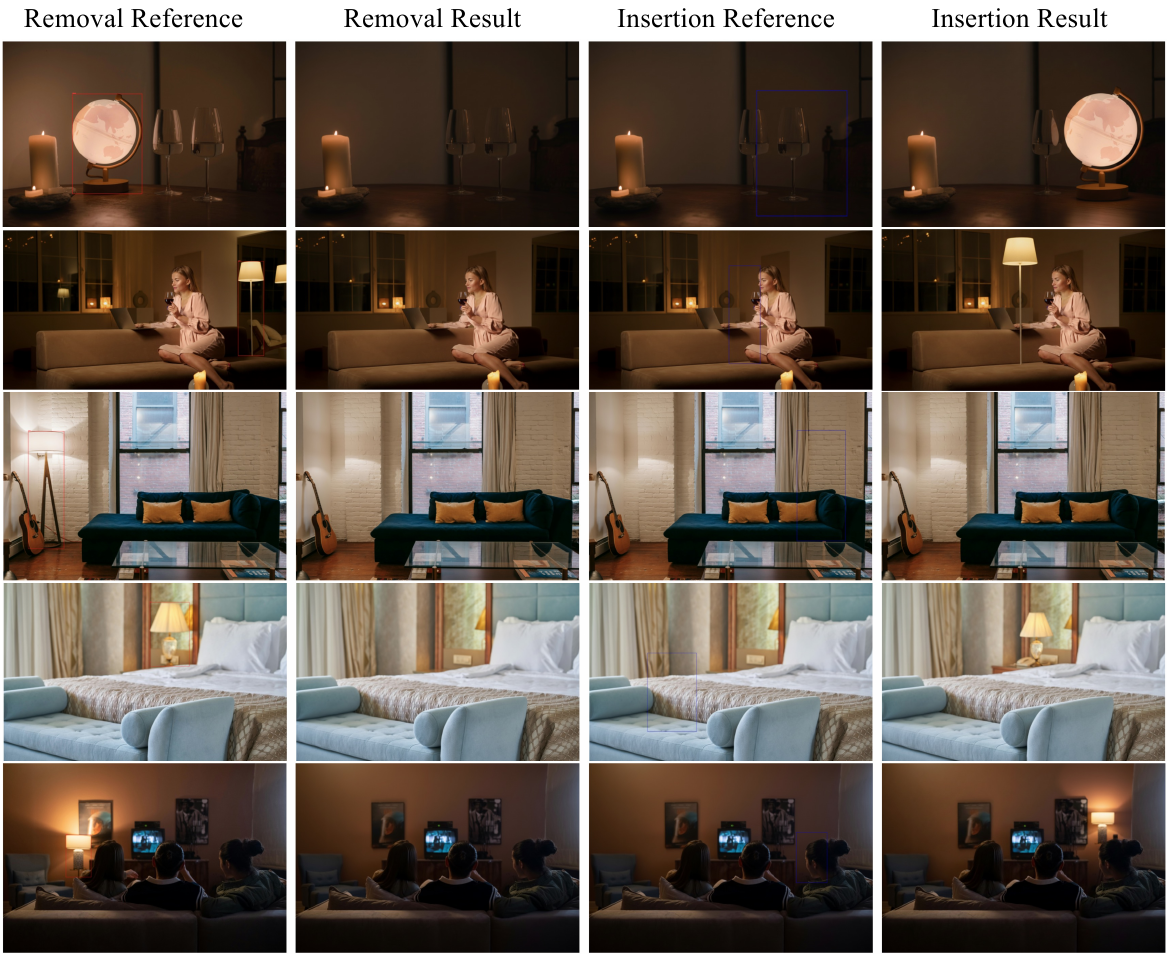}
    
    \caption{\textbf{Light movement results for Nano-Banana~\cite{Gemini2.5Flash2025} two-step editing.} We show the input reference image for removal and insertion tasks, and the results.}
    \label{figs:twostep}
\end{figure*}

For the two-step setup described in the main paper, we decompose the task into:  
(1) \emph{light removal}, followed by  
(2) \emph{light insertion at the target location}.  

We first segment the light-source object using SAM and crop it into a standalone patch. The model is then prompted twice: once to remove the light source from the scene, and once to reinsert it at the \textcolor{blue}{blue} target region with physically consistent relighting. We show the prompt for removal and insertion below.

\begin{figure}[h]
    \centering
\begin{tcolorbox}[
    colback=blue!5!white,
    colframe=blue!5!white,
    arc=2mm,
    boxrule=0pt,
    left=2mm, right=2mm, top=2mm, bottom=2mm,
    fontupper=\scriptsize
]
\textcolor{pink}{\{reference\_image\}}

\textcolor{RoyalBlue}{Remove the light source object located in the RED bounding box from the image.}

\textcolor{RoyalBlue}{Requirements:}

1. Fill the area naturally with appropriate background content.

2. Ensure the removal is seamless and photorealistic.

3. Update shadows and lighting to reflect the absence of the light source.
    \end{tcolorbox}
    \caption{Removal prompt for the two-step Nano-Banana editing.}
    \label{fig:onestep_basic}
\end{figure}

\begin{figure}[h]
    \centering
\begin{tcolorbox}[
    colback=blue!5!white,
    colframe=blue!5!white,
    arc=2mm,
    boxrule=0pt,
    left=2mm, right=2mm, top=2mm, bottom=2mm,
    fontupper=\scriptsize
]
\textcolor{pink}{\{object\_image\}}

\textcolor{pink}{\{reference\_image\}}

\textcolor{RoyalBlue}{You are provided with two images:}

1. A light source object to be inserted 

2. A scene image with a BLUE bounding box marking the target location\\
 
\textcolor{PineGreen}{Task: Insert the light source object from the first image into the BLUE bounding box location in the second image.}

\textcolor{RoyalBlue}{Requirements:}

- Place the object naturally within the blue box area

- Ensure seamless integration with the scene

- Add appropriate shadows and lighting effects from the new light source

- The result should be photorealistic with proper light-scene interactions

- Remove the blue bounding box in the final output

    \end{tcolorbox}
    \caption{Insertion prompt for the two-step Nano-Banana editing.}
    \label{fig:onestep_basic}
\end{figure}

We show the removal result and final editing results in Figure~\ref{figs:twostep}.

\section{More Qualitative Results}


\begin{figure*}[h]
    \centering
    \includegraphics[width=0.95\linewidth]{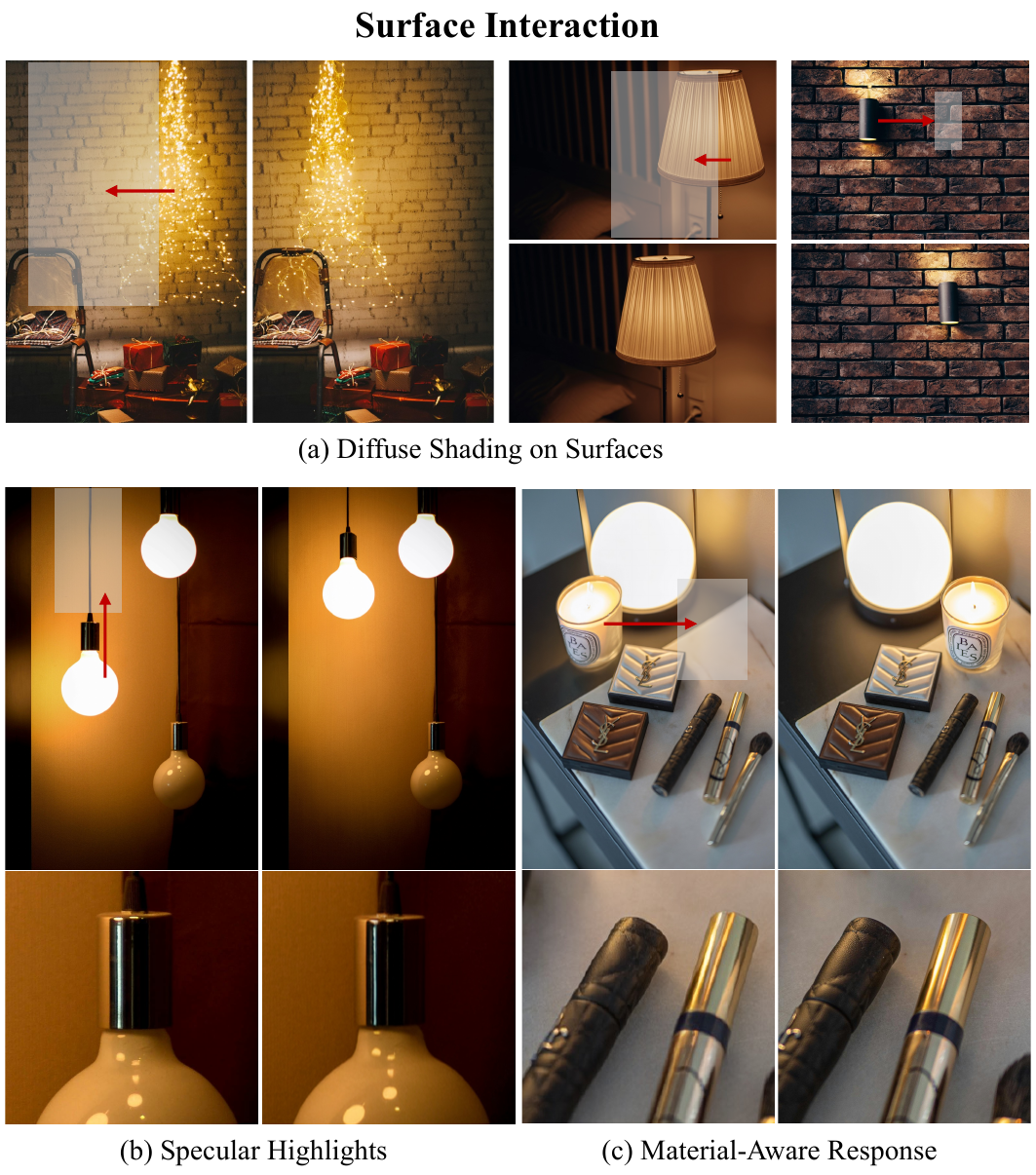}
    \caption{\textbf{Surface interaction results.} Examples of diffuse shading, specular highlights, and material-dependent responses under light movement.}
    \label{figs:surface}
\end{figure*}

\begin{figure*}[h]
    \centering
    \includegraphics[width=0.95\linewidth]{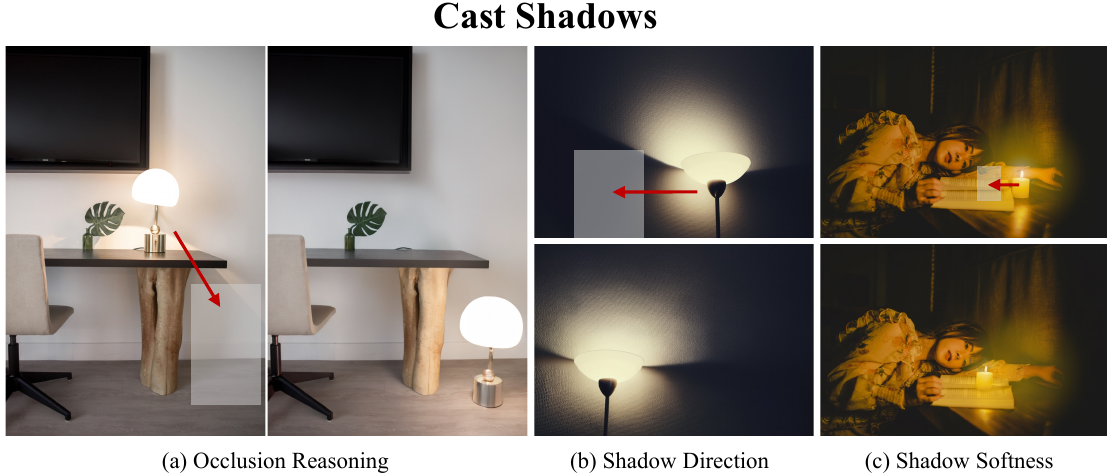}
    \caption{\textbf{Cast shadow results.} Examples demonstrating changes in shadow direction, softness, and occlusion structure.}
    \label{figs:shadow}
\end{figure*}

\begin{figure*}[h]
    \centering
    \includegraphics[width=0.95\linewidth]{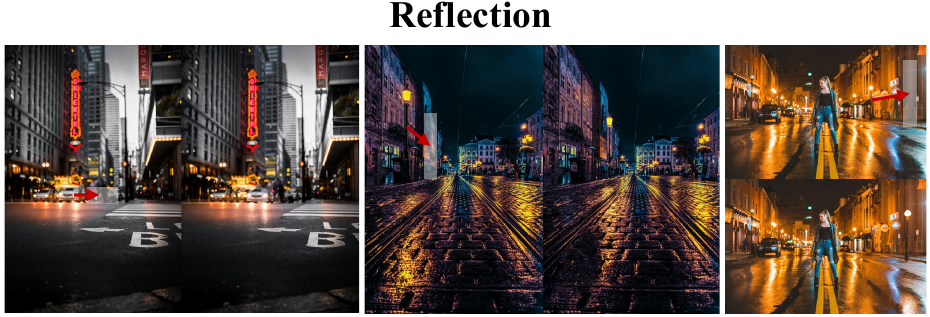}
    \caption{\textbf{Reflection results.} Examples of updated highlight positions and reflective consistency after repositioning the light source.}
    \label{figs:reflection}
\end{figure*}

To further analyze the behavior of our model under different illumination conditions, we present additional qualitative examples that highlight three key challenges in light movement: \emph{surface interaction}, \emph{cast shadows}, and \emph{reflections}. These examples illustrate how our method jointly reasons about geometry, materials, and global illumination when the light source is repositioned.

\subsection{Surface Interaction}

When a light source moves, surfaces must respond with updated shading patterns that reflect both geometry and material properties.  

\noindent\textbf{Diffuse Shading on Surfaces.} The overall shading gradients on diffuse surfaces shift according to light direction, requiring coherent large-scale brightness redistribution.  

\noindent\textbf{Specular Highlights.} Highlights must relocate precisely across curved or planar glossy surfaces, with intensity modulated by surface normals and microfacet properties.

\noindent\textbf{Material-Aware Response.} Different materials like metal, plastic, paper, and fabric exhibit distinct BRDF behaviors. \ours preserves these distinctions, producing sharp specular lobes on metallic surfaces and softer responses on matte objects.

The results in Figure~\ref{figs:surface} demonstrate that \ours accurately captures these interactions across diverse real-world scenes.

\subsection{Cast Shadows}

Shadows encode critical geometric information and are highly sensitive to light movement. This makes them one of the most challenging aspects of the task.  

\noindent\textbf{Shadow Direction \& Length.} As the light source moves, cast shadows must rotate and stretch according to object–surface geometry. A physically inconsistent direction immediately breaks realism.  

\noindent\textbf{Shadow Softness (Penumbra / Umbra).} The softness of a shadow depends on the light source size and distance. Our model adapts shadow edges to match these conditions, producing realistic penumbra transitions.  

\noindent\textbf{Occlusion Reasoning.} Correct shadow placement requires inferring hidden or partially visible geometry. Our method captures this structure, preserving accurate attachment points and occlusion boundaries.

As shown in Figure~\ref{figs:shadow}, the generated shadows are coherent, geometry-aware, and physically plausible.

\subsection{Reflections}

Reflections provide one of the strongest tests of illumination consistency since they depend jointly on light position, surface orientation, and scene geometry.  

\noindent\textbf{Reflection Direction Consistency.} Reflective surfaces: wet roads, metal, polished materials must display highlights and reflected objects at positions consistent with the moved light source.  

\noindent\textbf{Reflection Intensity and Falloff.} The brightness and attenuation of reflected highlights must adjust with the illumination energy and direction.  

\noindent\textbf{Environment \& Object Coupling.} Reflections must remain coherent with both the environment and the emitting light source, especially in urban scenes with multiple reflective cues.

Figure~\ref{figs:reflection} shows that \ours maintains this global consistency, accurately updating reflective behavior under light movement.

\section{Limitations}
\label{sec:limitations}

\ours currently targets single-image editing and may produce temporally inconsistent results when applied frame-by-frame to video; video extension is a natural direction for future work. The method struggles with large-scale ambient lighting changes (e.g., relocating outdoor sunlight) due to limited training coverage, and placement of lights behind transparent or heavily occluded objects remains challenging. The approach inherits the assumptions of 2D diffusion-based editing and does not recover explicit 3D geometry or materials, which can lead to physically implausible results in scenes with complex light transport.

\end{document}